\documentclass[10pt,twocolumn,letterpaper]{article}

\usepackage{cvpr}
\usepackage{times}
\usepackage{epsfig}
\usepackage{graphicx}
\usepackage{amsmath}
\usepackage{amssymb}
\usepackage{xcolor}
\usepackage{xspace}
\usepackage{multirow}
\usepackage{bbm}
\usepackage{multirow}
\usepackage{subcaption}
\usepackage{algorithm}
\usepackage[noend]{algpseudocode}
\usepackage{tabularx}
\usepackage{makecell}
\usepackage{microtype}
\usepackage{booktabs}
\usepackage[colorlinks=true,citecolor=blue,breaklinks=true,bookmarks=false]{hyperref}

\cvprfinalcopy 

\def\cvprPaperID{109} 

\newcommand{\colmap}{{\sc Colmap}\xspace}
\newcommand{\Dataset}{{MegaDepth}\xspace}
\newcommand{\DatasetShort}{{MD}\xspace}

\newcommand{\siMSE}{si-RMSE\xspace}
\newcommand{\sdr}{\text{SDR}\xspace}

\newcommand{\Ford}{F_\mathsf{ord}}
\newcommand{\Bord}{B_\mathsf{ord}}
\newcommand{\Lsi}{\mathcal{L}_\mathsf{si}}
\newcommand{\Ldata}{\mathcal{L}_\mathsf{data}}
\newcommand{\Lgrad}{\mathcal{L}_\mathsf{grad}}
\newcommand{\Lord}{\mathcal{L}_\mathsf{ord}}

\newcommand{\ord}{\operatorname{ord}}

\newenvironment{packed_enum}{
\begin{enumerate}
  \setlength{\itemsep}{1pt}
  \setlength{\parskip}{2pt}
  \setlength{\parsep}{0pt}
}{\end{enumerate}}

\newenvironment{packed_item}{
\begin{itemize}
  \setlength{\itemsep}{1pt}
  \setlength{\parskip}{2pt}
  \setlength{\parsep}{0pt}
}{\end{itemize}}

\makeatletter
\def\BState{\State\hskip-\ALG@thistlm}
\makeatother

\begin{document}

\title{MegaDepth: Learning Single-View Depth Prediction from 
  Internet Photos}
  
\author{Zhengqi Li \qquad Noah Snavely \\
Department of Computer Science $\&$ Cornell Tech, Cornell University 
}

\maketitle

\begin{abstract}
Single-view depth prediction is a fundamental problem in computer
vision. Recently, deep learning methods have led to significant
progress, but such methods are limited by the available training
data. Current datasets based on 3D sensors have key limitations,
including indoor-only images (NYU), small numbers of training examples
(Make3D), and sparse sampling (KITTI). We propose to use multi-view
Internet photo collections, a virtually unlimited data source, to
generate training data via modern structure-from-motion and multi-view
stereo (MVS) methods, and present a large depth dataset called
\Dataset based on this idea. Data derived from MVS comes with its own
challenges, including noise and unreconstructable objects.  We address
these challenges with new data cleaning methods, as well as
automatically augmenting our data with ordinal depth relations
generated using semantic segmentation. We validate the use of large
amounts of Internet data
by showing that models trained on \Dataset exhibit strong
generalization---not only to novel scenes, but also to other diverse
datasets including Make3D, KITTI, and DIW, even when no images from
those datasets are seen during training.\footnote{Project website: \url{http://www.cs.cornell.edu/projects/megadepth/}}

\end{abstract}

\section{Introduction}

\begin{figure}[t]
  \centering
    \begin{tabular}{@{\hspace{0.1em}}c@{\hspace{0.1em}}c@{\hspace{0.1em}}}
        \includegraphics[width=0.42\columnwidth]{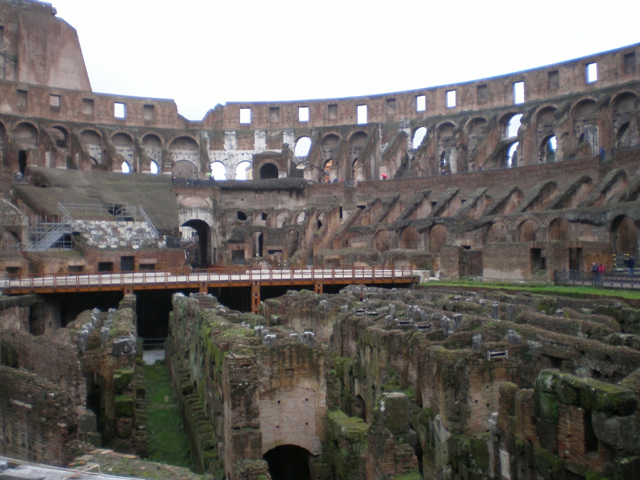} \vspace{-0.2em} & 
        \includegraphics[width=0.42\columnwidth]{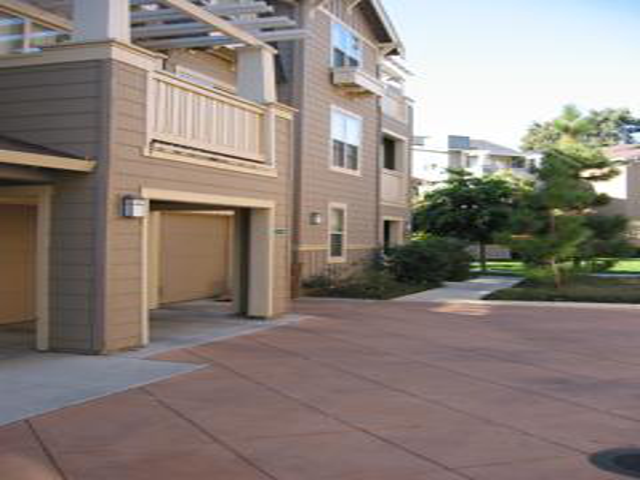}  \vspace{-0.1em} \\ 
      {\scriptsize (a) Internet photo of Colosseum} & {\scriptsize (b) Image from Make3D} \\  
        \includegraphics[width=0.42\columnwidth]{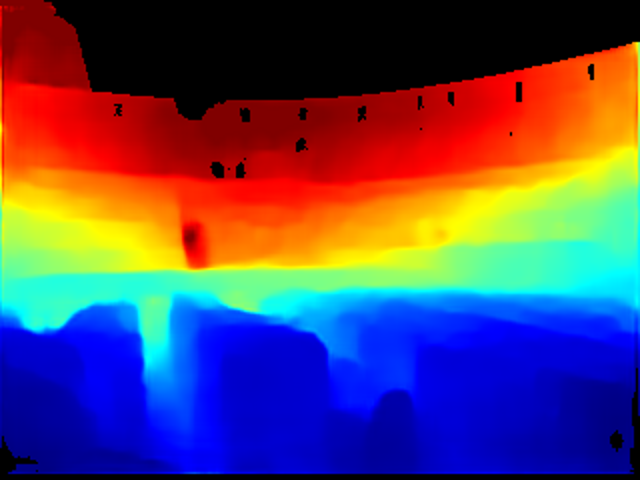} \vspace{-0.2em}  &
        \includegraphics[width=0.42\columnwidth]{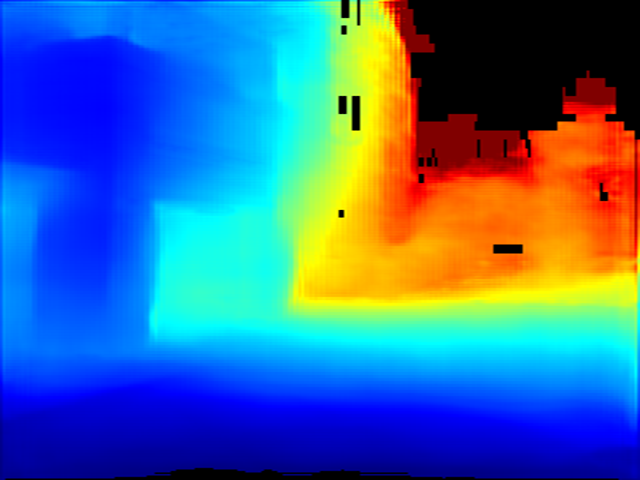}  \vspace{-0.1em} \\ 
         {\scriptsize (c) Our single-view depth prediction} & {\scriptsize (d) Our single-view depth prediction} \\  
   \end{tabular} 
    \centering
	\begin{tabular}{@{\hspace{0.1em}}c@{\hspace{0.1em}}}
        \includegraphics[width=0.90\columnwidth]{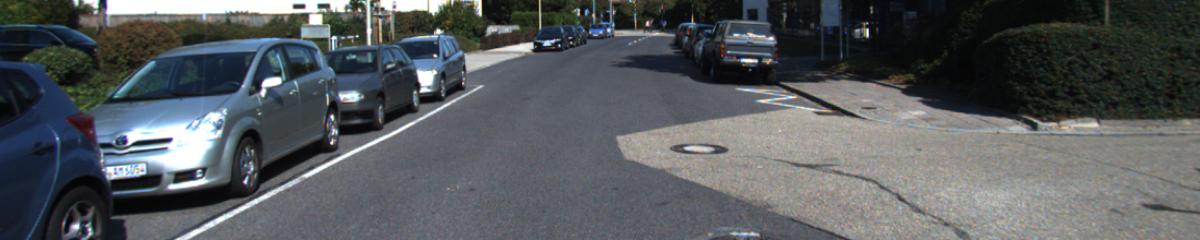} \vspace{-0.1em}  \vspace{-0.1em} \\ 
        {\scriptsize (e) Image from KITTI}  \\ 
        \includegraphics[width=0.90\columnwidth]{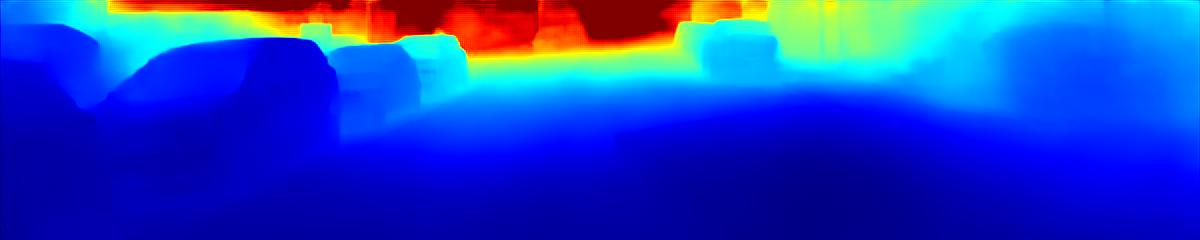} \vspace{-0.1em} \\
        {\scriptsize (f) Our single-view depth prediction} \vspace{-0.1em}
    \end{tabular} 
	   \caption{
       \label{fig:teaser} We use large Internet image
       collections, combined with 3D reconstruction and semantic
       labeling methods, to generate large amounts of training data
       for single-view depth prediction. {\bf (a), (b), (e):} Example
       input RGB images. {\bf (c), (d), (f):} Depth maps predicted by
       our \Dataset-trained CNN (blue$=$near, red$=$far). For these
       results, the network was not trained on Make3D and KITTI
       data.} \vspace{-0.5em}
\end{figure}

Predicting 3D shape from a single image is an important capability of
visual reasoning, with applications in robotics, graphics, and other
vision tasks such as intrinsic images.
While single-view depth estimation is a challenging, underconstrained
problem,
deep learning methods have recently driven significant progress. Such
methods thrive when trained with large amounts of data.
Unfortunately, fully general training data in the form of (\emph{RGB
image, depth map}) pairs is difficult to collect. Commodity RGB-D
sensors such as Kinect have been widely used for this
purpose~\cite{silberman2012indoor}, but are limited to indoor use.
Laser scanners
have enabled important datasets such as Make3D~\cite{saxena2009make3d}
and KITTI~\cite{Menze2015CVPR}, but such devices are cumbersome to
operate (in the case of industrial scanners), or produce sparse depth
maps (in the case of LIDAR). Moreover, both Make3D and KITTI
are collected in specific scenarios (a university campus, and atop a
car, respectively). Training data can also be generated through
crowdsourcing, but this approach has so far been limited to gathering
sparse ordinal relationships or surface
normals~\cite{gingold2012micro,chen2016single,Chen2017SurfaceNI}.

In this paper, we explore the use of a nearly unlimited source of data
for this problem: images from the Internet from overlapping
viewpoints, from which structure-from-motion (SfM) and multi-view
stereo (MVS) methods can automatically produce dense depth. Such
images
have been widely used in research on large-scale 3D
reconstruction~\cite{snavely2006photo,goesele2007multi,agarwal2009rome,frahm2010rome}.
We propose to use the outputs of these systems as the inputs to
machine learning methods for single-view depth prediction. By using
large amounts of diverse training data from photos taken around the
world, we seek to learn to predict depth with high accuracy and
generalizability. Based on this idea, we introduce \Dataset
(\DatasetShort), a large-scale depth dataset generated from Internet
photo collections, which we make fully available to the community.

To our knowledge, ours is the first use of Internet SfM+MVS data for
single-view depth prediction. Our main contribution is the
\DatasetShort dataset itself. In addition, in creating \DatasetShort,
we found that care must be taken in preparing a dataset from noisy MVS
data, and so we also propose new methods for processing raw MVS
output, and a corresponding new loss function for training models with
this data. Notably, because MVS tends to not reconstruct dynamic
objects (people, cars, etc), we augment our dataset with ordinal
depth relationships automatically derived from semantic segmentation,
and train with a joint loss that includes an ordinal term.
In our experiments, we show that by training on \DatasetShort, we can
learn a model that works well not only on images of new scenes,
but that also generalizes remarkably well to completely different
datasets, including Make3D, KITTI, and DIW---achieving
much better generalization than prior
datasets. Figure~\ref{fig:teaser}
shows example results spanning different test sets from a network
trained solely on our \DatasetShort dataset.

\section{Related work}\label{sec:related}
\noindent \textbf{Single-view depth prediction.} A variety of methods
have been proposed for single-view depth prediction, most recently by
utilizing machine
learning~\cite{hoiem2005geometric,saxena2005learning}.
A standard approach is to collect RGB images with ground truth depth,
and then train a model (e.g., a CNN) to predict depth from
RGB~\cite{eigen2014depth,Depth2015CVPR,Depth2015Liu,roy2016monocular,baig2016coupled,laina2016deeper}.
Most such methods are trained on a few standard datasets, such as
NYU~\cite{silberman2011indoor,silberman2012indoor}, Make3D
\cite{saxena2009make3d}, and KITTI~\cite{geiger2012kitti}, which are
captured using RGB-D sensors (such as Kinect) or laser scanning. Such
scanning methods have important limitations, as discussed in the
introduction. Recently, Novotny~\etal~\cite{novotny2017learning} trained
a network on 3D models derived from SfM+MVS on videos to learn 3D
shapes of single objects. However, their method is limited to
images of objects, rather than scenes.



Multiple views of a scene can also be used as an implicit source of
training data for single-view depth prediction, by utilizing view
synthesis
as a supervisory
signal~\cite{xie2016deep3d,garg2016unsupervised,monodepth17,zhou2017unsupervised}.
However, view synthesis is only a proxy for depth, and may not always
yield high-quality learned depth. Ummenhofer
\etal~\cite{ummenhofer2016demon}
trained from overlapping image pairs taken with a single camera, and
learned to predict image matches, camera poses, and depth. However, it
requires two input images at test time.


\smallskip
\noindent \textbf{Ordinal depth prediction.} Another way to collect
depth data for training is to ask people to manually annotate depth in
images. While labeling absolute depth is challenging, people are good
at specifying {\em relative} (ordinal) depth relationships (e.g., {\em
  closer-than}, {\em further-than})~\cite{gingold2012micro}.
Zoran \etal \cite{zoran2015learning} used such relative depth judgments
to predict ordinal relationships between points using CNNs.
Chen \etal leveraged crowdsourcing of ordinal depth labels to create a
large dataset called ``Depth in the
Wild''~\cite{chen2016single}. While useful for predicting depth
ordering (and so we incorporate ordinal data automatically generated
from our imagery), the Euclidean accuracy of depth learned solely from
ordinal data is limited.

\smallskip
\noindent \textbf{Depth estimation from Internet photos.}  Estimating
geometry from Internet photo collections has been an active research
area for a decade, with advances in both
structure from
motion~\cite{snavely2006photo,agarwal2009rome,Wu2013TowardsLI,schonberger2016structure}
and multi-view
stereo~\cite{goesele2007multi,furukawa2010towards,schonberger2016pixelwise}.
These techniques generally operate on 10s to 1000s of images. Using
such methods, past work has used retrieval and SfM to build a 3D model
seeded from a single image~\cite{schoenberger2015single}, or
registered a photo to an existing 3D model to transfer
depth~\cite{zhang2014personal}. However, this work requires either
having a detailed 3D model of each location in advance, or building
one at run-time.
Instead, we use SfM+MVS to
train a network that generalizes to novel locations and scenarios.



\section{The \Dataset Dataset}\label{sec:dataset}

In this section, we describe how we construct our dataset.
We first download Internet photos from Flickr for a set of
well-photographed landmarks from the Landmarks10K
dataset~\cite{li2016worldwide}.
We then reconstruct each landmark in 3D using state-of-the-art SfM and
MVS methods. This yields an SfM model as well as a dense depth map for
each reconstructed image. However, these depth maps have significant
noise and outliers, and training a deep network on this raw depth data
will not yield a useful predictor. Therefore, we propose a series
of processing steps that prepare these depth maps for use in learning,
and additionally use semantic segmentation to automatically generate
ordinal depth data.

\subsection{Photo calibration and reconstruction}
We build a 3D model from each photo collection using \colmap, a
state-of-art SfM system~\cite{schonberger2016structure} (for
reconstructing camera poses and sparse point clouds) and MVS
system~\cite{schonberger2016pixelwise} (for generating dense depth
maps). We use \colmap because we found that it produces high-quality
3D models via its careful incremental SfM procedure, but other such
systems could be used. \colmap produces a depth map $D$ for every
reconstructed photo $I$ (where some pixels of $D$ can be empty if
\colmap was unable to recover a depth), as well as other outputs, such
as camera parameters and sparse SfM points plus camera
visibility.



\subsection{Depth map refinement}\label{sec:clean_depth} 

\begin{figure}[t]
  \centering
    \begin{tabular}{@{\hspace{0.0em}}c@{\hspace{0.0em}}c@{\hspace{0.0em}}c@{\hspace{0.0em}}}
        \includegraphics[width=0.3\columnwidth]{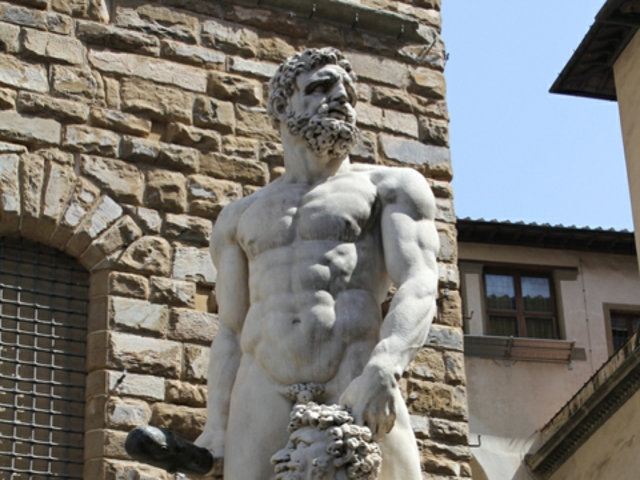} \vspace{-0.05em} & 
        \includegraphics[width=0.3\columnwidth]{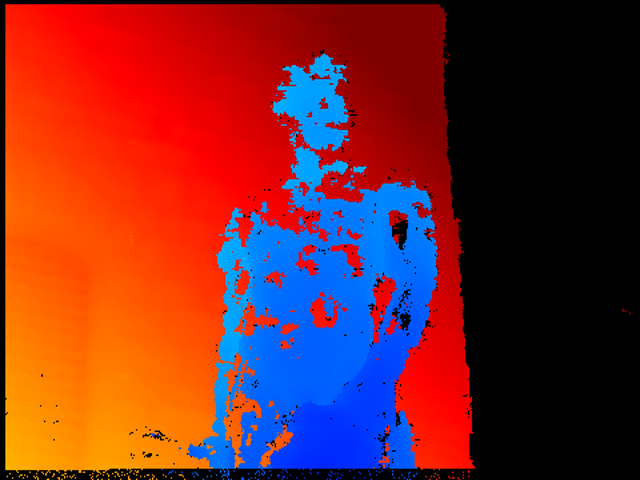}  \vspace{-0.05em} &
        \includegraphics[width=0.3\columnwidth]{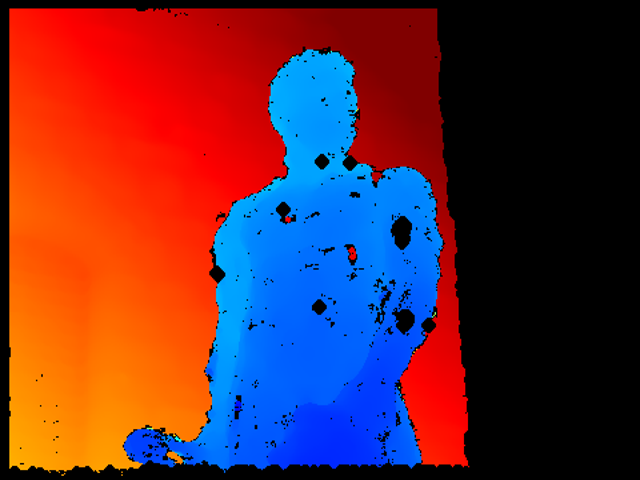} \vspace{-0.05em} \\   
        \includegraphics[width=0.3\columnwidth]{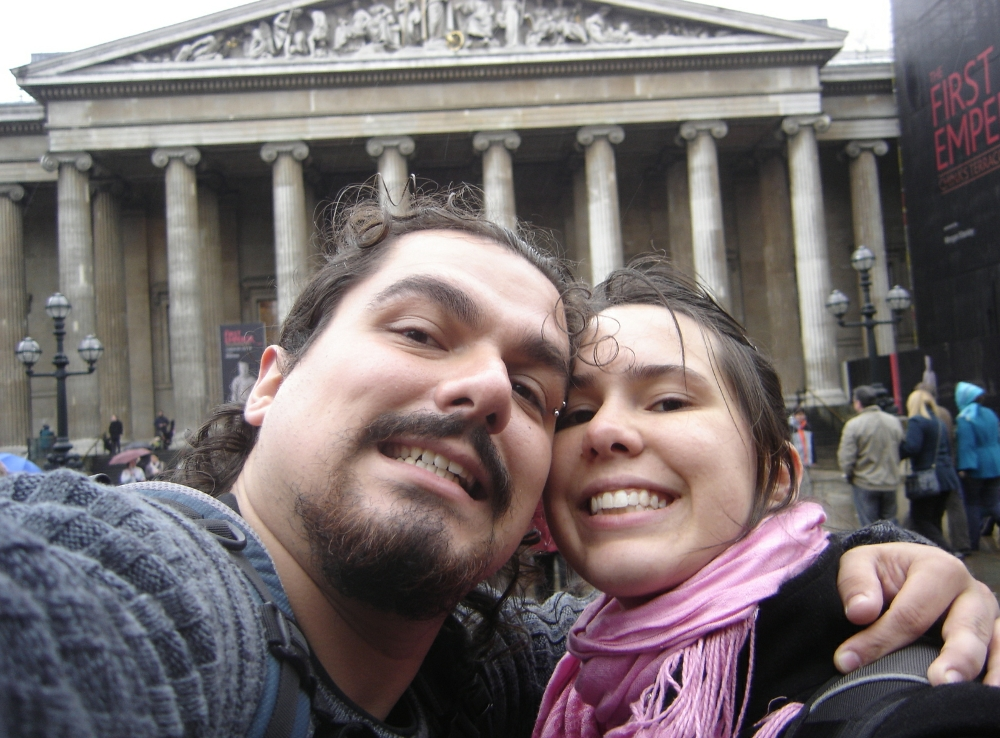} \vspace{0.0em} & 
        \includegraphics[width=0.3\columnwidth]{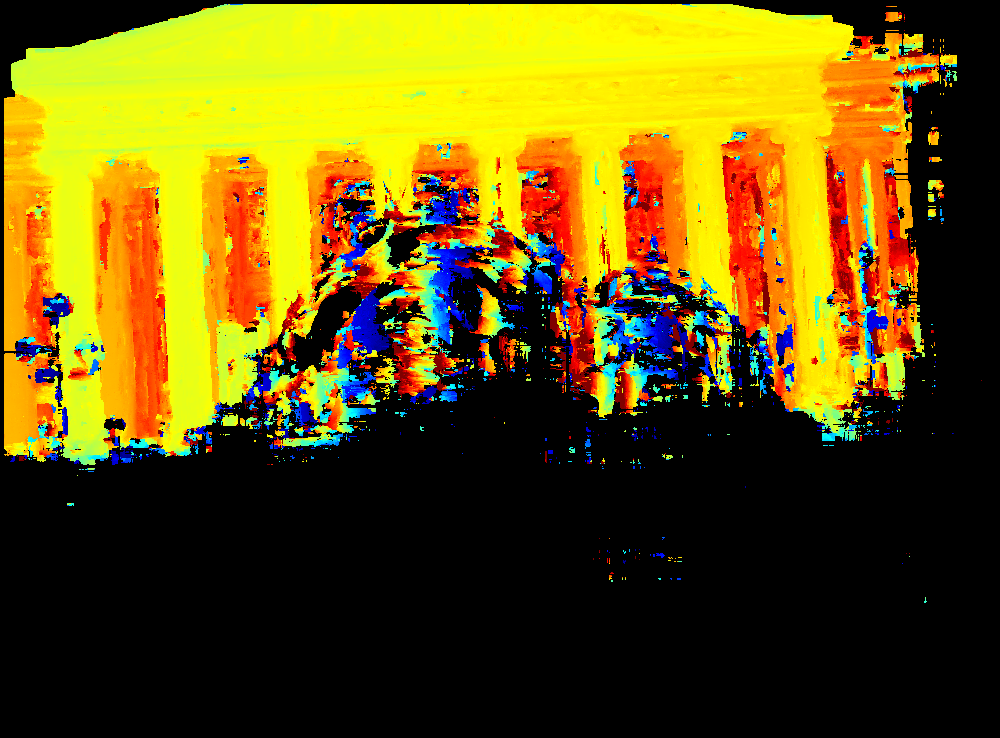}  \vspace{0.0em} &
        \includegraphics[width=0.3\columnwidth]{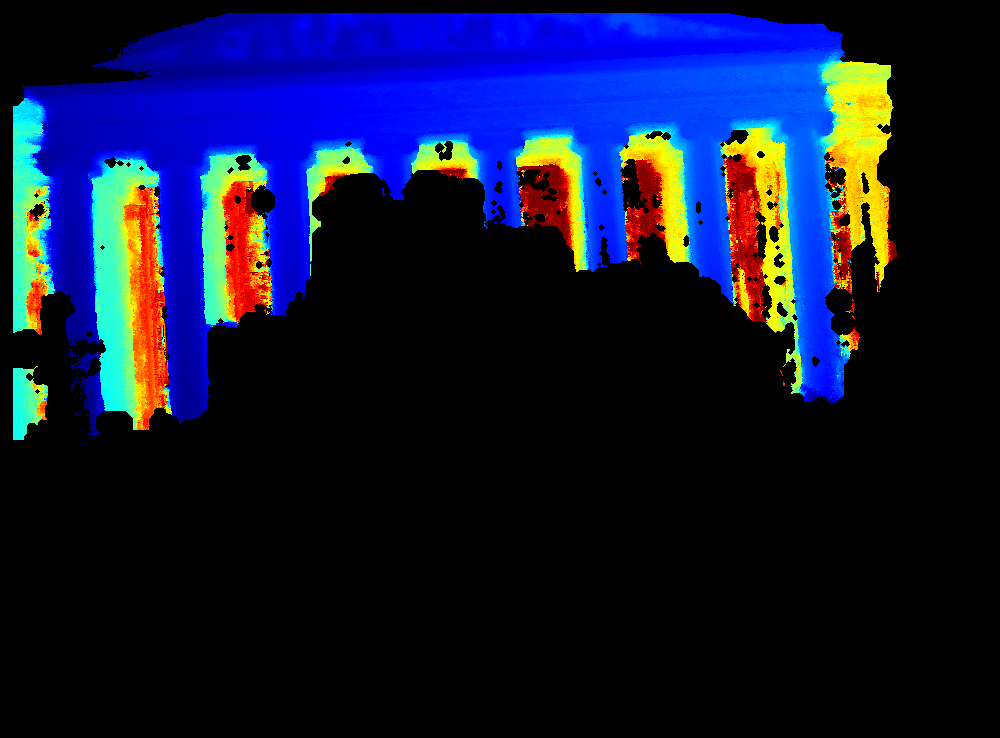} \vspace{0.0em} \\  
        {\small (a) Input photo} & {\small (b) Raw depth} & {\small
          (c) Refined depth} \vspace{-0.5em}
    \end{tabular} 
  \caption{\textbf{Comparison between MVS depth maps with and without
      our proposed refinement/cleaning methods.} The raw MVS
    depth maps (middle) exhibit depth bleeding (top) or incorrect
    depth on people (bottom). Our methods (right) can correct or
    remove such outlier depths.}\label{fig:clean_compare}
    \vspace{-1em}
\end{figure}

The raw depth maps from \colmap contain many outliers from a range of
sources, including: (1) transient objects (people, cars, etc.)  that
appear in a single image but nonetheless are assigned (incorrect)
depths, (2) noisy depth discontinuities,
and (3) bleeding of background depths into foreground objects.
Other MVS methods exhibit similar problems due to inherent ambiguities
in stereo matching. 
Figure~\ref{fig:clean_compare}(b) shows two example depth maps
produced by \colmap that illustrate these issues.
Such outliers have a highly negative effect on the depth prediction
networks we seek to train. To address this problem, we propose two new
depth refinement methods designed to generate high-quality training
data:

First, we devise a modified MVS algorithm based on \colmap, but more
conservative in its depth estimates, based on the idea that we would
prefer less training data over bad training data. \colmap computes
depth maps iteratively, at each stage trying to ensure geometric
consistency between nearby depth maps. One adverse effect of this
strategy is that background depths can tend to ``eat away'' at
foreground objects, because one way to increase consistency between
depth maps is to consistently predict the background depth (see
Figure~\ref{fig:clean_compare} (top)). To counter this effect, at each
depth inference iteration in \colmap, we compare the depth values at
each pixel before and after the update and keep the smaller (closer)
of the two.
We then apply a median filter to remove unstable depth values. We
describe our modified MVS algorithm in detail in the supplemental
material.

Second, we utilize semantic segmentation to enhance and filter the
depth maps, and to yield large amounts of ordinal depth comparisons as
additional training data. The second row of
Figure~\ref{fig:clean_compare} shows an example depth map computed
with our object-aware filtering.
We now describe our use of semantic segmentation
in detail.

\subsection{Depth enhancement via semantic segmentation}\label{sec:semantic} 

Multi-view stereo methods can have
problems with a number of object types, including transient objects
such as people and cars, difficult-to-reconstruct objects such as
poles and traffic signals, and sky regions. However, if we can
understand the semantic layout of an image, then we can attempt to
mitigate these issues, or at least identify problematic pixels. We
have found that deep learning methods for semantic segmentation are
starting to become reliable enough for this
use~\cite{zhao2016pyramid}.

We propose three new uses of semantic segmentation in the creation of
our
dataset. First, we use such segmentations to remove spurious MVS
depths in foreground regions. Second, we use the segmentation as a
criterion to categorize each photo as providing either Euclidean depth
or ordinal depth data. Finally, we combine semantic information and
MVS depth to automatically annotate ordinal depth relationships, which
can be used to help training in regions that cannot be reconstructed
by MVS.

\smallskip
\noindent \textbf{Semantic filtering.}  To process a given photo $I$,
we first run semantic segmentation using
PSPNet~\cite{zhao2016pyramid}, a recent segmentation method, trained
on the MIT Scene Parsing dataset (consisting of 150 semantic
categories)~\cite{zhou2017scene}. We then divide the pixels into
three
subsets by predicted semantic category:
\begin{packed_enum}
\item{\bf Foreground objects}, denoted $F$, corresponding to objects
  that often appear in the foreground of scenes, including static
  foreground objects (e.g., statues, fountains) and dynamic objects
  (e.g., people, cars).
\item{\bf Background objects}, denoted $B$, including buildings,
  towers, mountains, etc. (See supplemental material for full details
  of the foreground/background classes.)
\item{\bf Sky}, denoted $S$, which is treated as a special case in the
  depth filtering described below.
\end{packed_enum}
We use this semantic categorization of pixels in several ways. As
illustrated in Figure~\ref{fig:clean_compare} (bottom), transient
objects such as people can result in spurious
depths. To remove these from each image $I$, we
consider each connected component $C$ of the foreground mask $F$. If
$<50\%$ of pixels in $C$ have a reconstructed depth, we discard all
depths from $C$. We use a threshold of 50\%, rather than simply
removing all foreground depths, because pixels on certain objects in
$F$ (such as sculptures) can indeed be accurately reconstructed (and
we found that PSPNet can sometimes mistake sculptures and people for
one another). This simple filtering of foreground depths yields large
improvements in depth map quality.
Additionally, we remove reconstructed depths that fall inside the sky
region $S$, as such depths tend to be spurious.


\smallskip
\noindent \textbf{Euclidean vs.\ ordinal depth.} For each 3D model we
have thousands of reconstructed Internet photos, and ideally we would
use as much of this depth data as possible for training. However, some
depth maps are more reliable than others, due to factors such as the
accuracy of the estimated camera pose or the presence of large
occluders. Hence, we found that it is beneficial to limit training to
a subset of highly reliable depth maps.
We devise a simple but effective way to compute a subset of
high-quality depth maps, by thresholding by the fraction of
reconstructed pixels. In particular, if $\ge 30\%$ of an image $I$
(ignoring the sky region $S$) consists of valid depth values, then we
keep that image as training data for learning Euclidean depth.
This criterion
prefers images without large transient foreground objects (e.g., ``no
selfies''). At the same time, such foreground-heavy images are
extremely useful for another purpose: automatically generating
training data for learning {\em ordinal} depth relationships.

\begin{figure}[t]
  \centering
    \begin{tabular}{@{\hspace{-0.1em}}c@{\hspace{-0.1em}}c@{\hspace{-0.1em}}c@{\hspace{-0.1em}}c@{\hspace{-0.1em}}}
        \includegraphics[width=0.24\columnwidth]{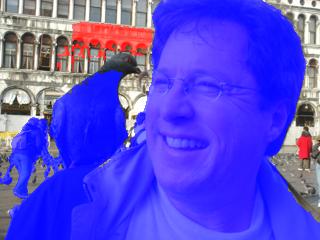} \vspace{-0.1em} & 
        \includegraphics[width=0.24\columnwidth]{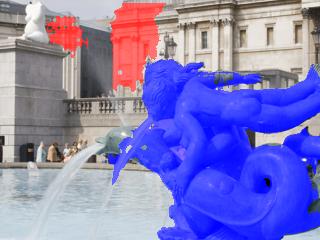}  \vspace{-0.1em} &
        \includegraphics[width=0.24\columnwidth]{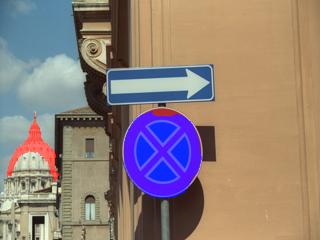} \vspace{-0.1em} &
        \includegraphics[width=0.24\columnwidth]{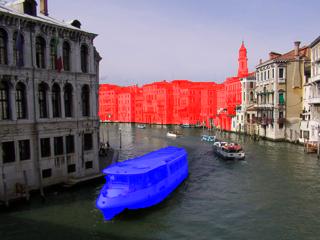} \vspace{-0.1em} \\
    \end{tabular} 
  \caption{ \textbf{Examples of automatic ordinal labeling.}
    \textcolor{blue}{Blue mask}: foreground ($\Ford$) derived from
    semantic segmentation. \textcolor{red}{Red mask}: background
    ($\Bord$) derived from reconstructed
    depth. \label{fig:automatic_label}} \vspace{-0.5em}
\end{figure}

\smallskip
\noindent \textbf{Automatic ordinal depth labeling.} As noted above,
transient or difficult to reconstruct objects, such as people, cars,
and street signs are often missing from MVS
reconstructions. Therefore, using Internet-derived data alone, we will
lack ground truth depth for such objects, and will likely do a poor
job of learning to reconstruct them.
To address this issue, we propose a novel method of automatically
extracting ordinal depth labels from our training images based on
their estimated 3D geometry and semantic segmentation.


Let us denote as $O$ (``Ordinal'') the subset of photos that do {\em
  not} satisfy the ``no selfies'' criterion described above. For each
image $I \in O$, we compute two regions, $\Ford \in F$ (based on
semantic information) and $\Bord \in B$ (based on 3D geometry
information), such that all pixels in $\Ford$ are likely closer to the
camera than 
all pixels in $\Bord$. Briefly, $\Ford$ consists of large connected
components of $F$, and $\Bord$ consists of large components of $B$
that also contain valid depths in the last quartile of the full depth
range for $I$ (see supplementary for full details).
We found this simple approach works very well ($>95\%$ accuracy in
pairwise ordinal relationships), likely because natural photos tend to
be composed in certain common ways.  Several examples of our automatic
ordinal depth labels are shown in Figure~\ref{fig:automatic_label}.

\subsection{Creating a dataset}
We use the approach above to densely reconstruct 200 3D models from
landmarks around the world, representing about 150K reconstructed
images. After our proposed filtering, we are left with 130K valid
images. Of these 130K photos, around 100K images are used for
Euclidean depth data, and the remaining 30K images are used to derive
ordinal depth data. We also include images from~\cite{Knapitsch2017}
in our training set. Together, this data comprises the \Dataset
(\DatasetShort) dataset, available at
\url{http://www.cs.cornell.edu/projects/megadepth/}.

\section{Depth estimation network}\label{sec:network}
This section presents our end-to-end deep learning algorithm for
predicting depth from a single photo.


\subsection{Network architecture}
We evaluated three
networks used in prior work on single-view depth prediction:
VGG~\cite{eigen2015predicting}, the ``hourglass''
network~\cite{chen2016single}, and a ResNet
architecture~\cite{laina2016deeper}. Of these, the hourglass network
performed best, as described in Section~\ref{sec:eval}.




\subsection{Loss function}
The 3D data produced by SfM+MVS is only up to an unknown scale
factor,
so we cannot compare predicted and ground truth depths directly. However, as noted by Eigen and Fergus~\cite{eigen2014depth},
the {\em ratios of pairs of depths} are preserved under scaling (or,
in the log-depth domain, the difference between pairs of
log-depths). Therefore, we solve for a depth map in the log domain
and train using a scale-invariant loss function, $\Lsi$. $\Lsi$
combines three terms:
\begin{align}
  \Lsi = \Ldata + \alpha \Lgrad + \beta \Lord. \label{eq:Loss}
\end{align}

\noindent \textbf{Scale-invariant data term.}  We adopt the loss of
Eigen and Fergus~\cite{eigen2014depth}, which computes the mean square
error (MSE) of the difference between \emph{all} pairs of log-depths
in linear time. Suppose we have a predicted log-depth map $L$, and a
ground truth log depth map $L^{*}$. $L_{i}$ and $L_{i}^{*}$ denote
corresponding individual log-depth values indexed by pixel position
$i$.  We denote $R_i = L_i - L_i^{*}$ and define:
\begin{align}
  \Ldata & = \frac{1}{n} \sum_{i=1}^n (R_i)^2  - \frac{1}{n^2} \left( \sum_{i=1}^{n}R_i \right)^2  \label{eq:data_term}
\end{align}    
%
%
%
%
%
where $n$ is the number of valid depths in the ground truth depth
map. 

\begin{figure}[t]
  \centering
    \begin{tabular}{@{\hspace{0.0em}}c@{\hspace{0.0em}}c@{\hspace{0.0em}}c@{\hspace{0.0em}}}
        \includegraphics[width=0.3\columnwidth]{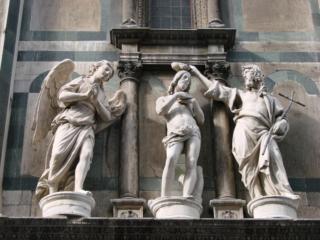} \vspace{-0.05em} & 
        \includegraphics[width=0.3\columnwidth]{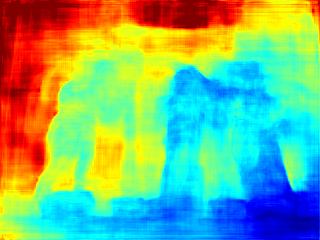}  \vspace{-0.05em} &
        \includegraphics[width=0.3\columnwidth]{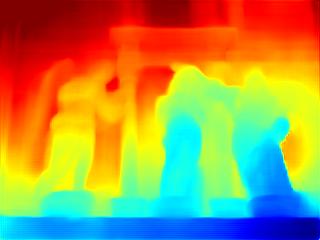} \vspace{-0.05em}\\   
        \includegraphics[width=0.3\columnwidth]{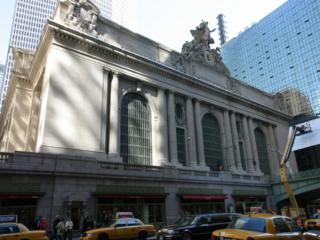} \vspace{-0.1em} &
        \includegraphics[width=0.3\columnwidth]{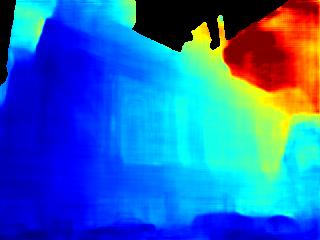} \vspace{-0.1em}&
        \includegraphics[width=0.3\columnwidth]{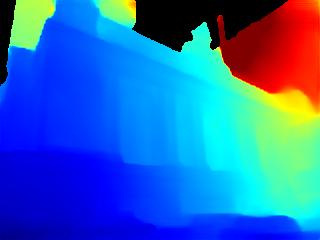} \vspace{-0.1em}\\
        {\scriptsize Input photo} & {\scriptsize Output w/o $\Lgrad$} & {\scriptsize Output w/ $\Lgrad$} \vspace{-0.5em}
    \end{tabular} 
  \caption{ \textbf{Effect of $\Lgrad$ term.}
    $\Lgrad$ encourages predictions to match the ground truth depth gradient.\label{fig:mgm}} \vspace{-0.5em}
\end{figure}

\smallskip
\noindent \textbf{Multi-scale scale-invariant gradient matching term.}
To encourage smoother gradient changes and sharper depth
discontinuities in the predicted depth map, we introduce a multi-scale
scale-invariant gradient matching term $\Lgrad$, defined as an
$\ell_1$ penalty on differences in log-depth gradients between the
predicted and ground truth depth map:
\begin{align}
  \Lgrad & = \frac{1}{n} \sum_{k} \sum_{i} \left( \left|\nabla_x R^k_i \right|
  + \left| \nabla_y R^k_i \right| \right) 
  \label{eq:smooth1}
\end{align}
where $R^k_i$ is the value of the log-depth difference map at position
$i$ and scale $k$. Because the loss is computed at multiple scales,
$\Lgrad$ captures depth gradients across large image distances. In our
experiments, we use four scales. We illustrate the effect of $\Lgrad$
in Figure~\ref{fig:mgm}.


\begin{figure}[t]
  \centering
    \begin{tabular}{@{\hspace{0.0em}}c@{\hspace{0.0em}}c@{\hspace{0.0em}}c@{\hspace{0.0em}}}
        \includegraphics[width=0.3\columnwidth]{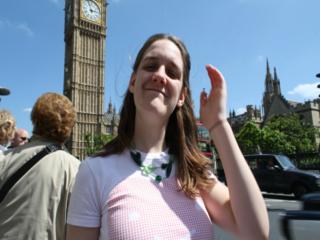} \vspace{-0.05em} & 
        \includegraphics[width=0.3\columnwidth]{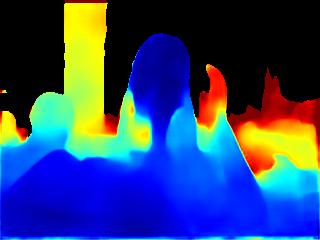}  \vspace{-0.05em} &
        \includegraphics[width=0.3\columnwidth]{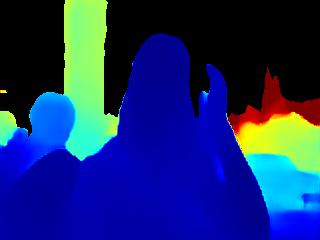} \vspace{-0.05em}\\   
        \includegraphics[width=0.3\columnwidth]{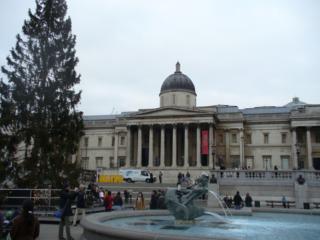} \vspace{-0.1em} &
        \includegraphics[width=0.3\columnwidth]{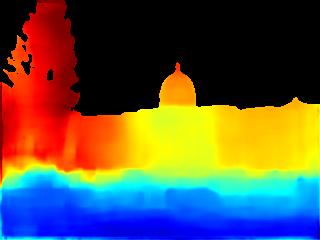} \vspace{-0.1em}&
        \includegraphics[width=0.3\columnwidth]{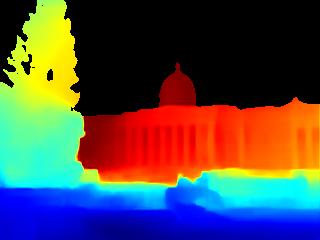} \vspace{-0.1em}\\
        {\scriptsize Input photo} & {\scriptsize Output w/o $\Lord$} & {\scriptsize Output w/ $\Lord$} \vspace{-0.5em}
    \end{tabular} 
  \caption{ \textbf{Effect of $\Lord$ term.} $\Lord$ tends to corrects
    ordinal depth relations for hard-to-construct objects such as the
    person in the first row and the tree in the second
    row.\label{fig:ordinal}} \vspace{-0.5em}
\end{figure}

\smallskip
\noindent \textbf{Robust ordinal depth loss.}  Inspired by
Chen~\etal~\cite{chen2016single}, our ordinal depth loss term $\Lord$
utilizes the automatic ordinal relations described in Section
\ref{sec:semantic}.
During training, for each image in our ordinal set $O$, we pick a
single pair of pixels $(i,j)$, with pixel $i$ and $j$ either belonging
to the foreground region $\Ford$ or the background region
$\Bord$. $\Lord$ is designed to be robust to the small number of
incorrectly ordered pairs.
\begin{align}
	\Lord = 
	   \begin{cases}
       \text{log} \left( 1+\text{exp} \left( P_{ij} \right) \right) &  \text{if }  P_{ij} \le \tau  \\ 
       \text{log} \left( 1+\text{exp} \left( \sqrt{P_{ij}} \right) \right) + c & \text{if }  P_{ij} > \tau
    \end{cases}
\end{align}
where $P_{ij} = - r_{ij}^{*} \left( L_i - L_j \right)$
and $r_{ij}^{*}$ is the automatically labeled ordinal depth relation
between $i$ and $j$ ($r_{ij}^{*} = 1$ if pixel $i$ is further than $j$
and $-1$ otherwise).
$c$ is a constant set so that $\Lord$ is continuous. 
$\Lord$ encourages the depth difference of a pair of points to be
large (and ordered) if our automatic labeling method judged the pair
to have a likely depth ordering.  We illustrate
the effect of $\Lord$ in Figure~\ref{fig:ordinal}. In our tests, we
set $\tau = 0.25$ based on cross-validation.



\section{Evaluation}\label{sec:eval}
In this section, we evaluate our networks on a number of datasets, and
compare to several state-of-art depth prediction algorithms, trained
on a variety of training data. In our evaluation, we seek to answer
several questions, including:
\begin{packed_item}
\item How well does our model trained on \DatasetShort generalize to
  new Internet photos from never-before-seen locations?
\item How important is our depth map processing? What is the effect of
  the terms in our loss function?
\item How well does our model trained on \DatasetShort generalize to
  other types of images from other datasets?
\end{packed_item}
The third question is perhaps the most interesting, because the
promise of training on large amounts of diverse data is good
generalization.
Therefore, we run a set of experiments training on one dataset and
testing on another, and show that our \DatasetShort dataset gives the
best generalization performance.

We also show that
our depth refinement strategies are essential for achieving good
generalization, and show that our proposed loss function---combining
scale-invariant data terms with an ordinal depth loss---improves
prediction performance
both quantitatively and qualitatively.

\smallskip
\noindent \textbf{Experimental setup.}  Out of the 200 reconstructed
models in our \DatasetShort dataset, we randomly select 46 to form a
test set (locations not seen during training).
For the remaining 154 models, we randomly split images from each
model into training and validation sets with a ratio of
96\% and 4\% respectively. We set $\alpha = 0.5$ and $\beta = 0.1 $ using \DatasetShort validation set.
We implement our networks in PyTorch~\cite{pytorch}, and train
using Adam~\cite{Kingma2014AdamAM} for 20 epochs with batch size 32. 

For fair comparison, we train and validate our network using
\DatasetShort data for all experiments. Due to variance in performance
of cross-dataset testing, we train four models on \DatasetShort and
compute the average error (see supplemental material for the
performance of each individual model).

\subsection{Evaluation and ablation study on \DatasetShort test set}
In this subsection, we describe experiments where we train on our
\DatasetShort training set and test on the \DatasetShort test set.

\smallskip
\noindent{\bf Error metrics.}  For numerical evaluation, we use two
scale-invariant error measures (as with our loss function, we use
scale-invariant measures due to the scale-free nature of SfM
models). The first measure is the scale-invariant RMSE (\siMSE)
(Equation~\ref{eq:data_term}), which measures precise numerical depth
accuracy.
The second measure is based on the preservation of depth ordering.
In particular, we use a measure similar to \cite{zoran2015learning,chen2016single}
that we call the {\em SfM Disagreement Rate} (\sdr).  \sdr is based on
the rate of disagreement with ordinal depth relationships derived from
estimated SfM points. We use sparse SfM points rather than dense MVS
because we found that sparse SfM points capture some structures not
reconstructed by MVS (e.g., complex objects such as lampposts).
We define $\sdr(D,D^*)$, the ordinal disagreement rate between the
predicted (non-log) depth map $D = \exp(L)$ and ground-truth SfM
depths $D^*$, as:
\begin{equation}
\resizebox{.9\columnwidth}{!} 
{
  $\sdr(D,D^*) = \frac{1}{n} \sum_{i,j\in\mathcal{P}}
  \mathbbm{1}\left(\ord(D_i,D_j) \neq \ord(D^{*}_i,D^{*}_j)\right)$
}
\end{equation}
where $\mathcal{P}$ is the set of pairs of pixels with available SfM
depths to compare, $n$ is the total number of pairwise comparisons,
and $\ord(\cdot,\cdot)$ is one of three depth relations ({\em
  further-than}, {\em closer-than}, and {\em same-depth-as}):
{\small
\begin{align}
	\ord(D_i,D_j) = 
	   \begin{cases}
      1 & \text{if}\ \frac{D_i}{D_j} > 1 + \delta \\ -1 &
      \text{if}\ \frac{D_i}{D_j} < 1 - \delta \\ 0 & \text{if}\ 1 -
      \delta \leq \frac{D_i}{D_j} \leq 1 + \delta
    \end{cases}
\end{align}
}
We also define $\sdr^{=}$ and $\sdr^{\neq}$ as the disagreement rate
with $\ord(D^{*}_i,D^{*}_j)=0$ and $\ord(D^{*}_i,D^{*}_j)\neq 0$
respectively. In our experiments, we set $\delta = 0.1$ for tolerance
to uncertainty in SfM points. For efficiency, we sample SfM points
from the full set to compute this error term.

\begin{table}[t]
\centering
{\small
\begin{tabular}{lrrrr}
 \toprule
Network & \siMSE & $\sdr^{=}$\% & $\sdr^{\neq}$\% & $\sdr$\% \\
\midrule
$\text{VGG}^\ast$~\cite{eigen2015predicting}  & 0.116 & 31.28 & 28.63 & 29.78 \\
VGG (full) & 0.115 & 29.64 & 27.22 & 28.40 \\
ResNet (full) & 0.124 & \textbf{27.32} & 25.35 & 26.27 \\
HG (full)  & \textbf{0.104} & 27.73 & \textbf{24.36} & \textbf{25.82} \\
\bottomrule
\end{tabular}
}
\caption{{\bf Results on the \DatasetShort test set (places unseen
    during training) for several network architectures.} For
  $\text{VGG}^\ast$ we use the same loss and network architecture as
  in~\cite{eigen2015predicting} for comparison
  to~\cite{eigen2015predicting}. Lower is better.\label{tb:tb1}}
\end{table}

\begin{table}[t]
\centering
{\small
\begin{tabular}{lrrrr}
  \toprule
Method & \siMSE  & $\sdr^{=}$\% & $\sdr^{\neq}$\% & $\sdr$\% 
 \\
\midrule
$\Ldata$ only      & 0.148 & 33.20 & 30.65 & 31.75 \\
\ +$\Lgrad$        & 0.123 & \textbf{26.17} & 28.32  & 27.11 \\
\ +$\Lgrad$ +$\Lord$  & \textbf{0.104} & 27.73 & \textbf{24.36} & \textbf{25.82} \\
\bottomrule
\end{tabular}
}
\caption{{\bf Results on \DatasetShort test set (places unseen during
    training) for different loss configurations.} Lower is
  better. \label{tb:tb2}}
\vspace{-1.0em}
\end{table}

\begin{table}[t]
\centering
{\small
\begin{tabular}{llrr}
\toprule
Test set & Error measure & Raw \DatasetShort & Clean \DatasetShort
\\
\midrule
Make3D & RMS & 11.41 & \textbf{5.322}\\
       & Abs Rel & 0.614 & \textbf{0.364}\\
       & log10 & 0.386 & \textbf{0.152}\\
\midrule
KITTI  & RMS & 12.15 & \textbf{6.680} \\
       & RMS(log) & 0.582 & \textbf{0.414}\\
       & Abs Rel & 0.433 & \textbf{0.368}\\
       & Sq Rel & 3.927 & \textbf{2.587}\\
\midrule
DIW & WHDR\% & 31.32 & \textbf{24.55} \\
\bottomrule
\end{tabular}
}
\caption{{\bf Results on three different test sets with and without
    our depth refinement methods}. {\em Raw \DatasetShort} indicates
  raw depth data; {\em Clean \DatasetShort} indicates depth data using
  our refinement methods. Lower is better for all error measures. \label{tb:tb3}}
\end{table}

\begin{figure*}[ptb]
 \centering
\centering
    \begin{subfigure}[b]{0.12\textwidth}
        \includegraphics[width=\textwidth]{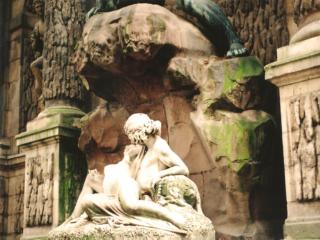}
    \end{subfigure} \hspace*{-0.8em}
    ~
    \begin{subfigure}[b]{0.12\textwidth}
        \includegraphics[width=\textwidth]{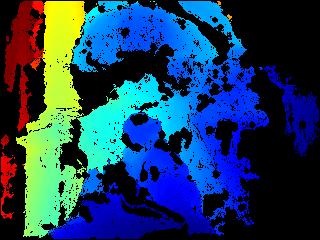}
    \end{subfigure} \hspace*{-0.8em}
    ~
    \begin{subfigure}[b]{0.12\textwidth}
        \includegraphics[width=\textwidth]{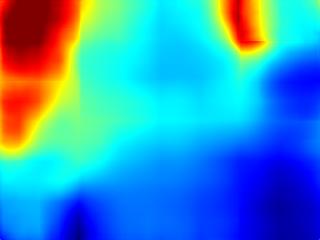}
    \end{subfigure} \hspace*{-0.8em}
    ~
    \begin{subfigure}[b]{0.12\textwidth}
        \includegraphics[width=\textwidth]{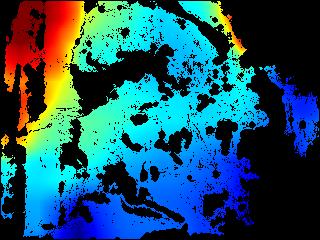}
    \end{subfigure} \hspace*{-0.8em}
    ~    
    \begin{subfigure}[b]{0.12\textwidth}
        \includegraphics[width=\textwidth]{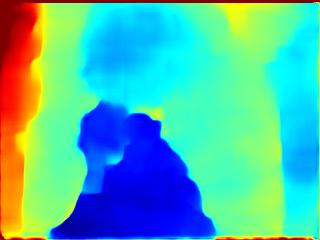}
    \end{subfigure}   \hspace*{-0.8em}
    ~
    \begin{subfigure}[b]{0.12\textwidth}
        \includegraphics[width=\textwidth]{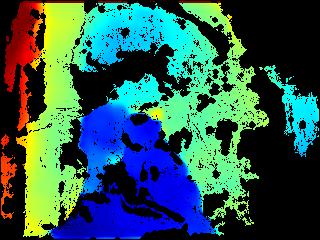}
    \end{subfigure}   \hspace*{-0.8em}
    ~    
    \begin{subfigure}[b]{0.12\textwidth}
        \includegraphics[width=\textwidth]{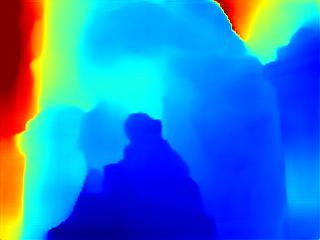}
    \end{subfigure} \hspace*{-0.8em}
    ~    
    \begin{subfigure}[b]{0.12\textwidth}
        \includegraphics[width=\textwidth]{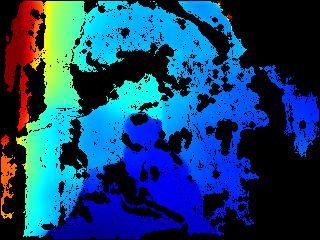}
    \end{subfigure}  
	 \begin{subfigure}[b]{0.12\textwidth}
        \includegraphics[width=\textwidth]{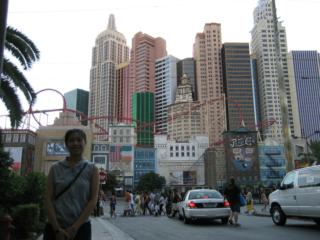}
    \end{subfigure} \hspace*{-0.8em}
    ~
    \begin{subfigure}[b]{0.12\textwidth}
        \includegraphics[width=\textwidth]{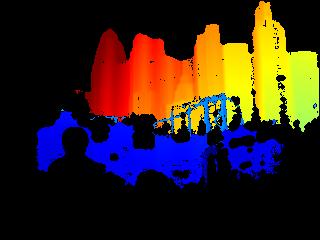}
    \end{subfigure} \hspace*{-0.8em}
    ~
    \begin{subfigure}[b]{0.12\textwidth}
        \includegraphics[width=\textwidth]{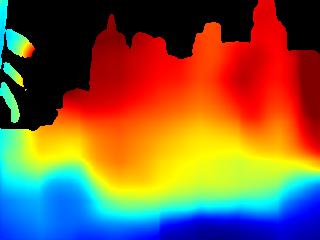}
    \end{subfigure} \hspace*{-0.8em}
    ~
    \begin{subfigure}[b]{0.12\textwidth}
        \includegraphics[width=\textwidth]{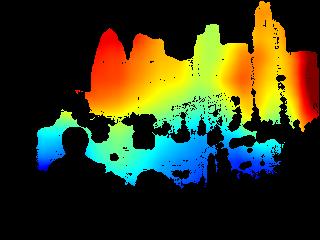}
    \end{subfigure} \hspace*{-0.8em}
    ~    
    \begin{subfigure}[b]{0.12\textwidth}
        \includegraphics[width=\textwidth]{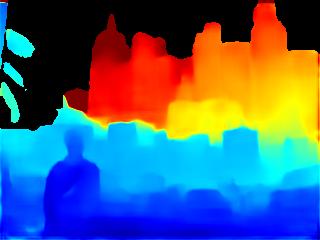}
    \end{subfigure}   \hspace*{-0.8em}
    ~
    \begin{subfigure}[b]{0.12\textwidth}
        \includegraphics[width=\textwidth]{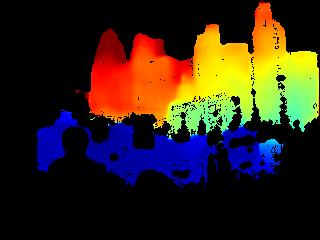}
    \end{subfigure}   \hspace*{-0.8em}
    ~    
    \begin{subfigure}[b]{0.12\textwidth}
        \includegraphics[width=\textwidth]{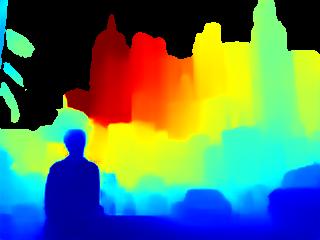}
    \end{subfigure} \hspace*{-0.8em}
    ~    
    \begin{subfigure}[b]{0.12\textwidth}
        \includegraphics[width=\textwidth]{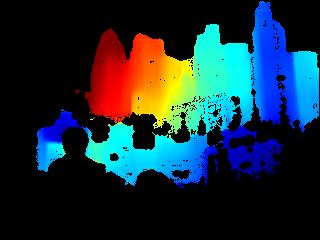}
    \end{subfigure}  
	 \begin{subfigure}[b]{0.12\textwidth}
        \includegraphics[width=\textwidth]{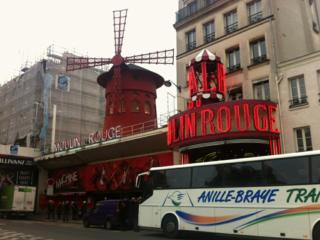}
    \end{subfigure} \hspace*{-0.8em}
    ~
    \begin{subfigure}[b]{0.12\textwidth}
        \includegraphics[width=\textwidth]{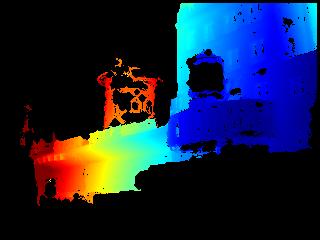}
    \end{subfigure} \hspace*{-0.8em}
    ~
    \begin{subfigure}[b]{0.12\textwidth}
        \includegraphics[width=\textwidth]{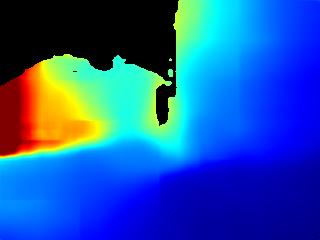}
    \end{subfigure} \hspace*{-0.8em}
    ~
    \begin{subfigure}[b]{0.12\textwidth}
        \includegraphics[width=\textwidth]{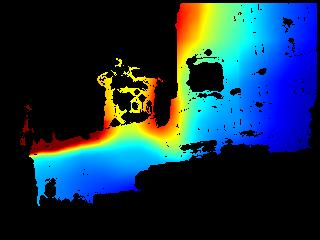}
    \end{subfigure} \hspace*{-0.8em}
    ~    
    \begin{subfigure}[b]{0.12\textwidth}
        \includegraphics[width=\textwidth]{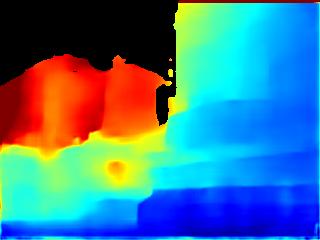}
    \end{subfigure}   \hspace*{-0.8em}
    ~
    \begin{subfigure}[b]{0.12\textwidth}
        \includegraphics[width=\textwidth]{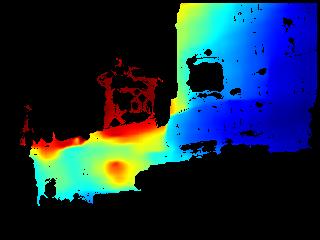}
    \end{subfigure}   \hspace*{-0.8em}
    ~    
    \begin{subfigure}[b]{0.12\textwidth}
        \includegraphics[width=\textwidth]{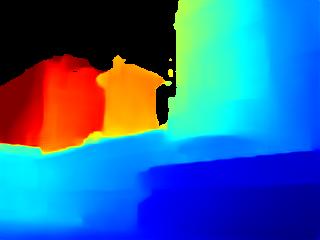}
    \end{subfigure} \hspace*{-0.8em}
    ~    
    \begin{subfigure}[b]{0.12\textwidth}
        \includegraphics[width=\textwidth]{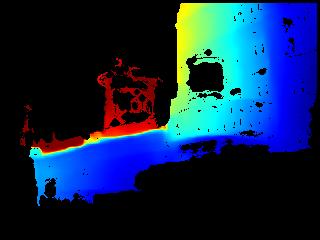}
    \end{subfigure}       
	 \begin{subfigure}[b]{0.12\textwidth}
        \includegraphics[width=\textwidth]{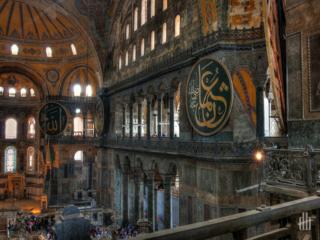}
		\caption{Image}   
    \end{subfigure} \hspace*{-0.8em}
    ~
    \begin{subfigure}[b]{0.12\textwidth}
        \includegraphics[width=\textwidth]{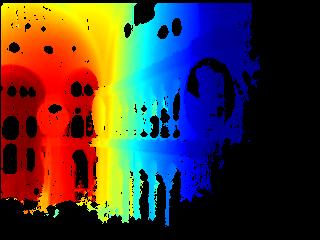}
		\caption{GT}   
    \end{subfigure} \hspace*{-0.8em}
    ~
    \begin{subfigure}[b]{0.12\textwidth}
        \includegraphics[width=\textwidth]{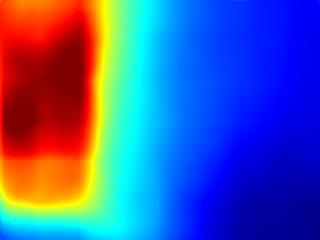}
		\caption{$\text{VGG}^{\ast}$}
    \end{subfigure} \hspace*{-0.8em}
    ~
    \begin{subfigure}[b]{0.12\textwidth}
        \includegraphics[width=\textwidth]{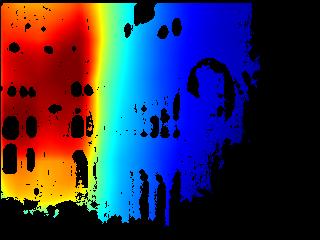}
        \caption{$\text{VGG}^{\ast}$ (M)}
    \end{subfigure} \hspace*{-0.8em}
    ~    
    \begin{subfigure}[b]{0.12\textwidth}
        \includegraphics[width=\textwidth]{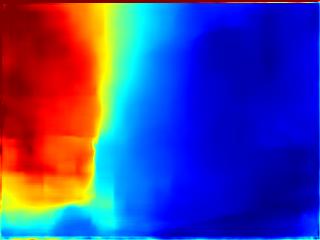}
        \caption{ResNet}
    \end{subfigure}   \hspace*{-0.8em}
    ~
    \begin{subfigure}[b]{0.12\textwidth}
        \includegraphics[width=\textwidth]{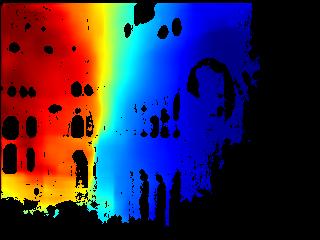}
        \caption{ResNet (M)}
    \end{subfigure}   \hspace*{-0.8em}
    ~    
    \begin{subfigure}[b]{0.12\textwidth}
        \includegraphics[width=\textwidth]{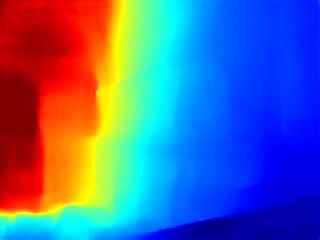}
        \caption{HG}
    \end{subfigure} \hspace*{-0.8em}
    ~    
    \begin{subfigure}[b]{0.12\textwidth}
        \includegraphics[width=\textwidth]{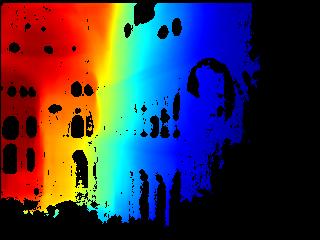}
        \caption{HG (M)} 
    \end{subfigure}  
    \caption{\textbf{Depth predictions on \DatasetShort test set.}
      (Blue$=$near, red$=$far.) For visualization, we mask out the
      detected sky region.  In the columns marked (M), we apply the
      mask from the GT depth map (indicating valid reconstructed
      depths) to the prediction map, to aid comparison with GT. (a)
      Input photo. (b) Ground truth \colmap depth map (GT).
      $\text{VGG}^{\ast}$ prediction using the loss and network
      of~\cite{eigen2015predicting}.  (d) GT-masked version of
      (c). (e) Depth prediction from a ResNet~\cite{laina2016deeper}. (f) GT-masked version of
      (e). (g) Depth prediction from an hourglass (HG)
      network~\cite{chen2016single} . (h) GT-masked version of
      (g).\label{fig:visual_compare}}
\end{figure*}

\smallskip
\noindent{\bf Effect of network and loss variants.}
We evaluate three popular network architectures for depth prediction
on our \DatasetShort test set: the VGG network used by Eigen
\etal~\cite{eigen2015predicting}, an ``hourglass''(HG)
network~\cite{chen2016single}, and ResNets~\cite{laina2016deeper}. To
compare our loss function to that of Eigen
\etal~\cite{eigen2015predicting}, we also test the same network and
loss function as \cite{eigen2015predicting} trained on \DatasetShort.
\cite{eigen2015predicting} uses a VGG network with a scale-invariant
loss plus single scale gradient matching term. Quantitative results
are shown in Table~\ref{tb:tb1} and qualitative comparisons are shown
in Figure~\ref{fig:visual_compare}.
We also evaluate variants of our method
trained using only some of our loss terms: (1) a version with only the
scale-invariant data term $\Ldata$ (the same loss as
in~\cite{eigen2014depth}),
(2) a version that adds our multi-scale gradient matching loss
$\Lgrad$, and (3) the full version including $\Lgrad$ and the ordinal
depth loss $\Lord$. Results are shown in Table~\ref{tb:tb2}.

As shown in Tables~\ref{tb:tb1} and \ref{tb:tb2}, the HG architecture
achieves the best performance of the three architectures,
and training with our full loss yields better
performance compared to other loss variants, including that of
\cite{eigen2015predicting} (first row of
Table~\ref{tb:tb1}). One thing to notice that is adding $\Lord$ could significantly improve $\sdr^{\neq}$ while increasing $\sdr^{=}$. 
Figure~\ref{fig:visual_compare} shows that our
joint loss helps preserve the structure of the depth map and capture
nearby objects such as people and buses.




Finally, we experiment with training our network on \DatasetShort with
and without our proposed depth refinement methods, testing on three
datasets: KITTI, Make3D, and DIW. The results, shown in
Table~\ref{tb:tb3}, show that networks trained on raw MVS depth do not
generalize well.
Our proposed refinements significantly boost prediction performance.





\subsection{Generalization to other datasets}
A powerful application of our 3D-reconstruction-derived training data
is to generalize to outdoor images beyond landmark photos. To evaluate
this capability, we train our model on \DatasetShort and test on three
standard benchmarks: Make3D ~\cite{saxena2005learning}, KITTI
~\cite{geiger2012kitti}, and DIW
~\cite{chen2016single}---\emph{without} seeing training data from
these datasets. Since our depth prediction is defined up to a scale
factor, for each dataset, we align each prediction from all non-target dataset trained models with the ground truth by a scalar computed from least sqaure solution to the ratio between ground truth and predicted depth.

\begin{table}[t]
\centering
{\small
\resizebox{\columnwidth}{!}{%
\begin{tabular}{llrrr}
  \toprule
Training set & Method & RMS & Abs Rel & log10 
 \\
\midrule
Make3D & Karsch~\etal~\cite{karsch2012depth} & 9.20 & 0.355 & 0.127  \\
 & Liu~\etal~\cite{liu2014discrete} & 9.49 & 0.335 & 0.137  \\
 & Liu~\etal~\cite{Depth2015CVPR} & 8.60 & 0.314 & 0.119  \\
 & Li~\etal~\cite{li2015depth} & 7.19 & 0.278 & 0.092 \\
 & Laina~\etal~\cite{laina2016deeper} & 4.45 & 0.176 & 0.072 \\
 & Xu~\etal~\cite{xu2017multi} & 4.38 & 0.184 & 0.065 \\
\midrule
NYU & Eigen~\etal~\cite{eigen2015predicting} & 6.89 & 0.505 & 0.198 \\
 & Liu~\etal~\cite{Depth2015CVPR} & 7.20 & 0.669 & 0.212 \\
 & Laina~\etal~\cite{laina2016deeper}  & 7.31 & 0.669 & 0.216 \\
\midrule
KITTI & Zhou~\etal~\cite{zhou2017unsupervised} & 8.39 & 0.651 & 0.231 \\
 & Godard~\etal~\cite{monodepth17} & 9.88 & 0.525 & 0.319 \\
\midrule
DIW & Chen~\etal~\cite{chen2016single} & 7.25 & 0.550  & 0.200 \\
\midrule
\DatasetShort & Ours & 6.23 & 0.402 & 0.156 \\
{\DatasetShort}+Make3D & Ours & 4.25 & 0.178 & 0.064 \\
\bottomrule
\end{tabular}
}
}

\caption{{\bf Results on Make3D for various training datasets and
    methods.} The first column indicates the training dataset. Errors
  for ``Ours'' are averaged over four models trained/validated on
  \DatasetShort. Lower is better for all metrics. \label{tb:tb4}}
\vspace{-0.5em}
\end{table}


\begin{figure}[t]
  \centering
    \begin{tabular}{@{\hspace{-0.1em}}c@{\hspace{-0.1em}}c@{\hspace{-0.1em}}c@{\hspace{-0.1em}}c@{\hspace{-0.1em}}c@{\hspace{-0.1em}}c@{\hspace{-0.1em}}}
        \includegraphics[width=0.16\columnwidth]{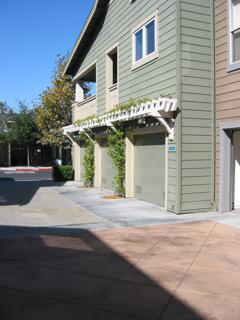} \vspace{-0.1em} & 
        \includegraphics[width=0.16\columnwidth]{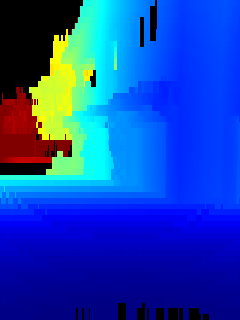}  \vspace{-0.1em} &
        \includegraphics[width=0.16\columnwidth]{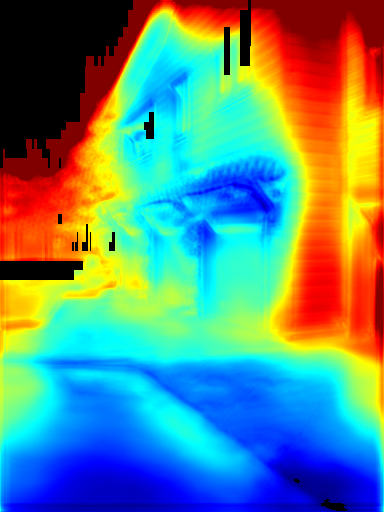}  \vspace{-0.1em} &
        \includegraphics[width=0.16\columnwidth]{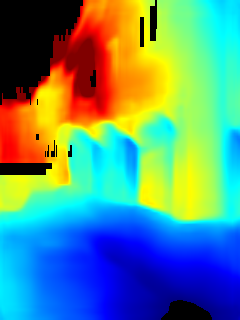}  \vspace{-0.1em} &
        \includegraphics[width=0.16\columnwidth]{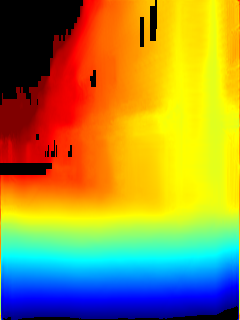}  \vspace{-0.1em} &
        \includegraphics[width=0.16\columnwidth]{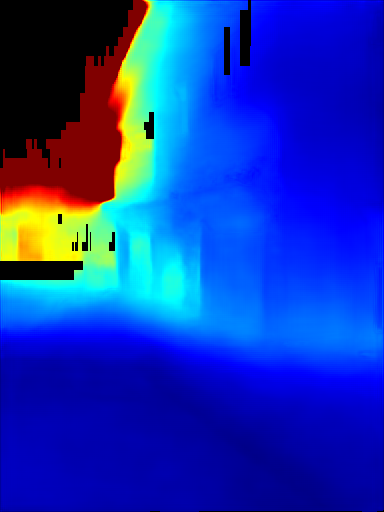} \vspace{-0.1em}\\
        \includegraphics[width=0.16\columnwidth]{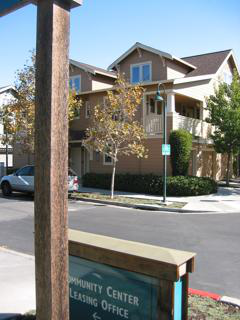} \vspace{-0.1em} & 
      \includegraphics[width=0.16\columnwidth]{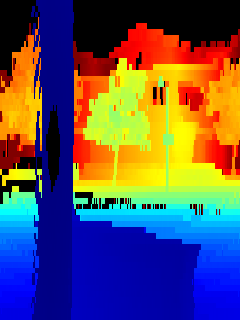}  \vspace{-0.1em} &
     \includegraphics[width=0.16\columnwidth]{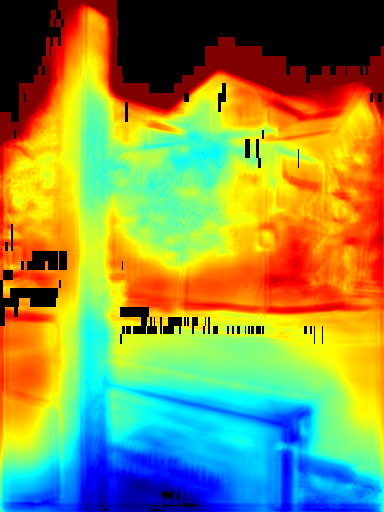}  \vspace{-0.1em} &
      \includegraphics[width=0.16\columnwidth]{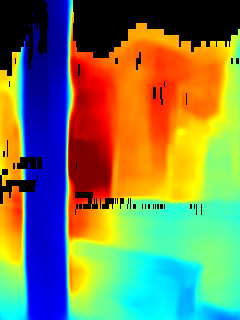}  \vspace{-0.1em} &
      \includegraphics[width=0.16\columnwidth]{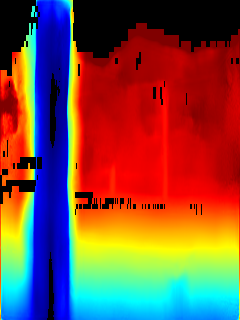}  \vspace{-0.1em} &
      \includegraphics[width=0.16\columnwidth]{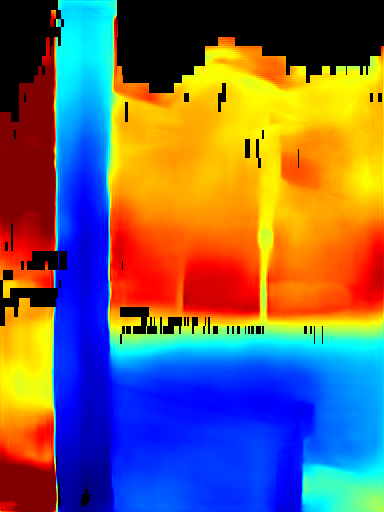} \vspace{-0.1em} \\
     
        \includegraphics[width=0.16\columnwidth]{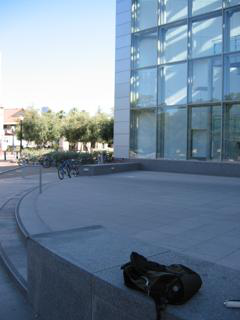} \vspace{-0.1em} & 
      \includegraphics[width=0.16\columnwidth]{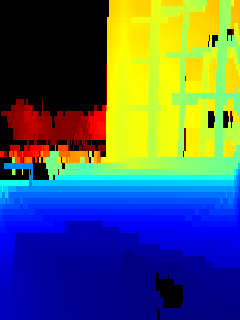}  \vspace{-0.1em} &
      \includegraphics[width=0.16\columnwidth]{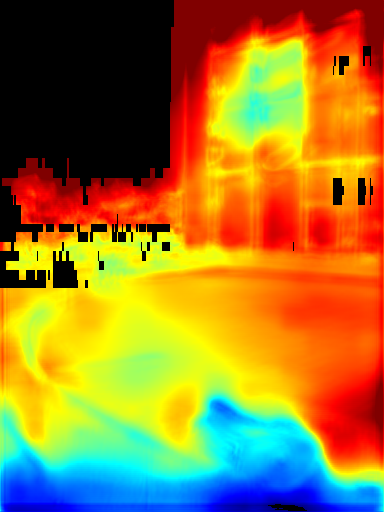}  \vspace{-0.1em} &
      \includegraphics[width=0.16\columnwidth]{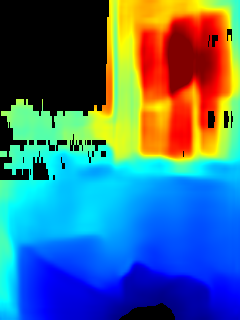}  \vspace{-0.1em} &
      \includegraphics[width=0.16\columnwidth]{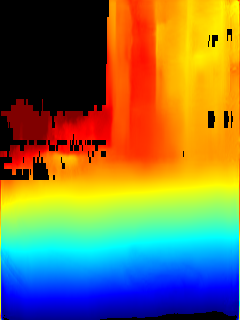}  \vspace{-0.1em} &
      \includegraphics[width=0.16\columnwidth]{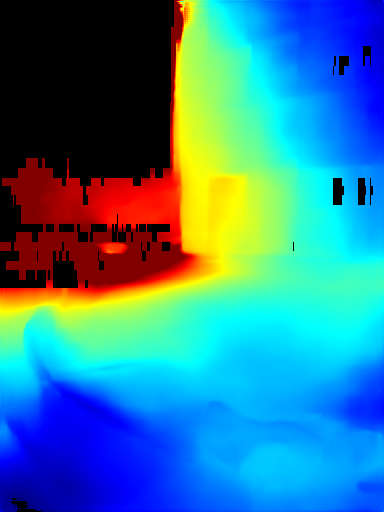} \vspace{-0.1em} \\        
        {\scriptsize (a) Image} & {\scriptsize (b) GT} & {\scriptsize (c) DIW~\cite{chen2016single}} & {\scriptsize (d) NYU~\cite{eigen2015predicting}} & {\scriptsize (e) KITTI~\cite{monodepth17}} & {\scriptsize (f) \DatasetShort} \vspace{-0.5em}
    \end{tabular} 
    \caption{ \textbf{Depth predictions on Make3D.}
      The last four columns show results from the best models trained
      on non-Make3D datasets (final column is our
      result).\label{fig:Make3d}} \vspace{-0.5em}
\end{figure}

\begin{table}[t]
\centering
{\small
  \resizebox{\columnwidth}{!}{%
    \begin{tabular}{llrrrr}
      \toprule
Training set & Method & RMS & RMS(log) & Abs Rel & Sq Rel \\
\midrule
KITTI & Liu~\etal~\cite{Depth2015Liu} & 6.52 & 0.275 & 0.202 & 1.614 \\
 & Eigen~\etal~\cite{eigen2014depth} & 6.31 & 0.282 & 0.203 & 1.548  \\
 & Zhou~\etal~\cite{zhou2017unsupervised} & 6.86 & 0.283 & 0.208 & 1.768  \\
 & Godard~\etal~\cite{monodepth17} & 5.93 & 0.247 & 0.148 & 1.334 \\
\midrule
Make3D & Laina~\etal~\cite{laina2016deeper} & 8.68 & 0.422 & 0.339 & 3.136 \\
 & Liu~\etal~\cite{Depth2015CVPR} & 8.70 & 0.447 & 0.362 & 3.465  \\
\midrule
NYU & Eigen~\etal~\cite{eigen2015predicting} & 10.37 & 0.510 & 0.521 & 5.016  \\
 & Liu~\etal~\cite{Depth2015CVPR} & 10.10 & 0.526 & 0.540 & 5.059 \\
 & Laina~\etal~\cite{laina2016deeper} & 10.07 & 0.527 & 0.515 & 5.049 \\
\midrule
CS & Zhou~\etal~\cite{zhou2017unsupervised} & 7.58 & 0.334 & 0.267 & 2.686 \\
\midrule
DIW & Chen~\etal~\cite{chen2016single} & 7.12 & 0.474 & 0.393 & 3.260 \\
\midrule
\DatasetShort & Ours & 6.68 & 0.414 & 0.368 & 2.587 \\
{\DatasetShort}+KITTI & Ours & 5.25 & 0.229 & 0.139 & 1.325 \\
\bottomrule
\end{tabular}
  }
}
\caption{{\bf Results on the KITTI test set for various training
    datasets and approaches.} Columns are as in
  Table~\ref{tb:tb4}.\label{tb:tb5}}
\vspace{-0.5em}
\end{table}

\begin{figure*}[btp]
\centering
\begin{tabular}{@{\hspace{-0.1em}}c@{\hspace{-0.1em}}c@{\hspace{-0.1em}}c@{\hspace{-0.1em}}c@{\hspace{-0.1em}}c@{\hspace{-0.1em}}c@{\hspace{-0.1em}}}
 \includegraphics[width=0.16\textwidth]{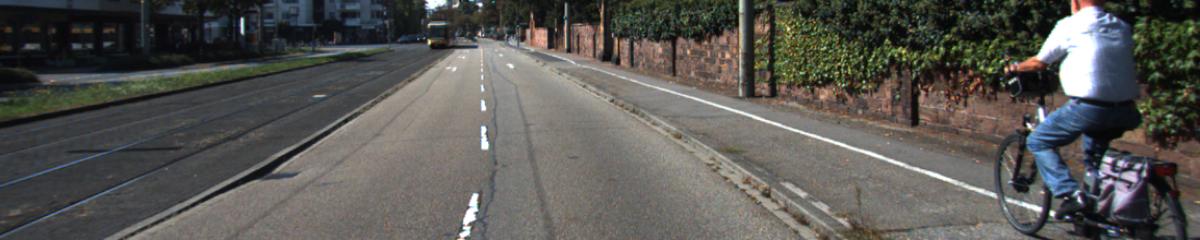} \vspace{-0.03em}&
 \includegraphics[width=0.16\textwidth]{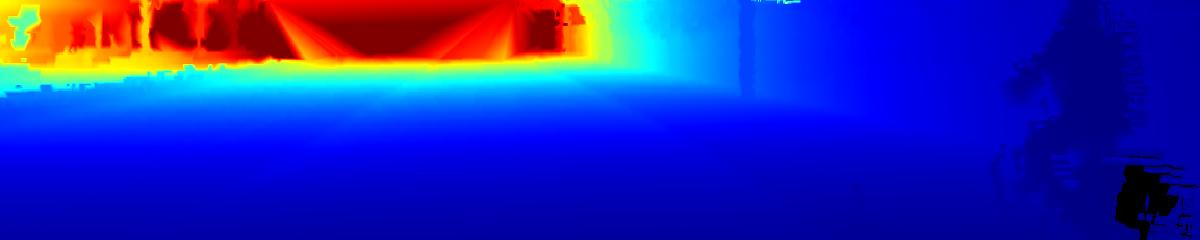} \vspace{-0.03em}&
 \includegraphics[width=0.16\textwidth]{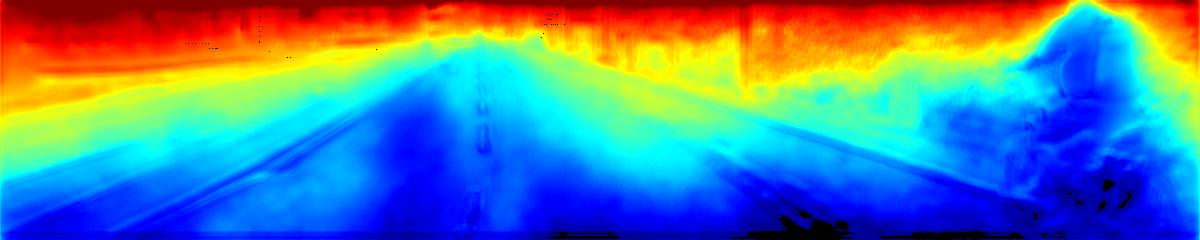} \vspace{-0.03em}&
 \includegraphics[width=0.16\textwidth]{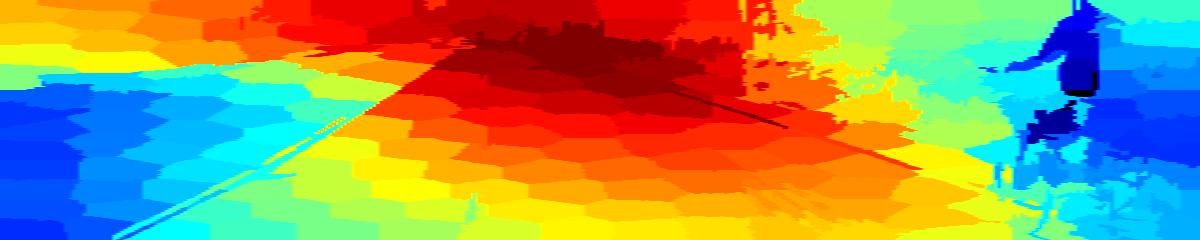} \vspace{-0.03em}&
 \includegraphics[width=0.16\textwidth]{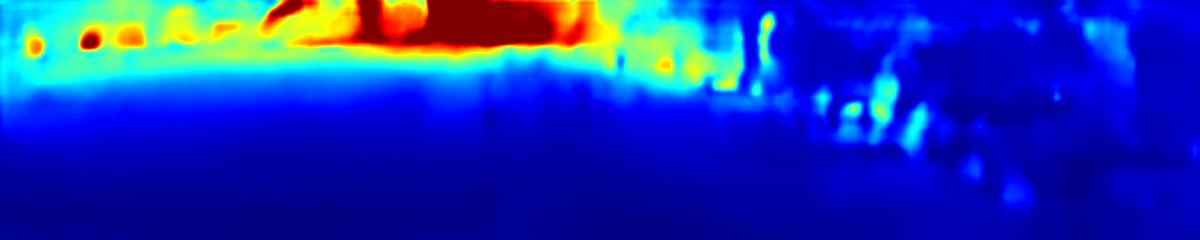} \vspace{-0.03em} &
 \includegraphics[width=0.16\textwidth]{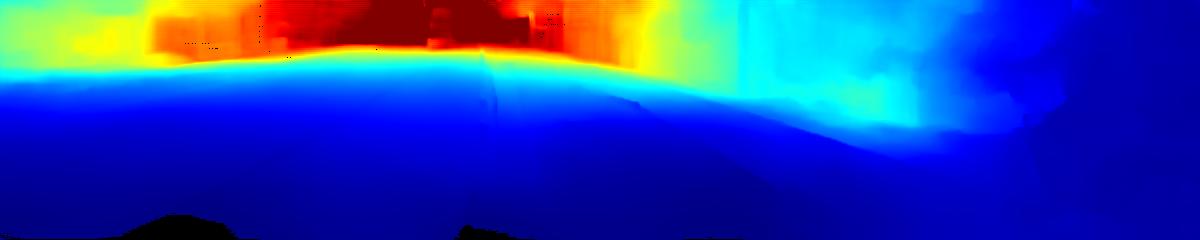} \\
  \includegraphics[width=0.16\textwidth]{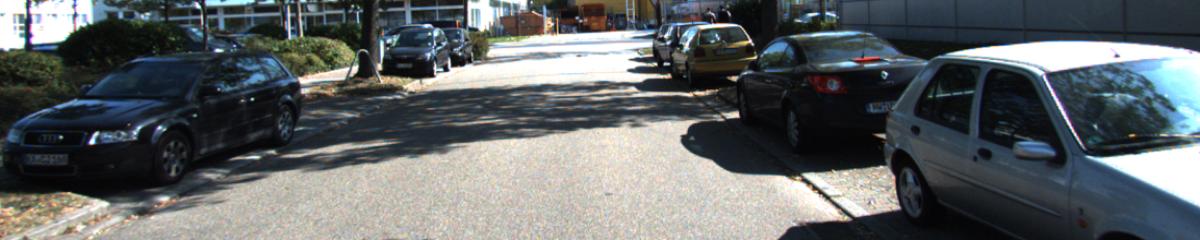} \vspace{-0.03em} &
  \includegraphics[width=0.16\textwidth]{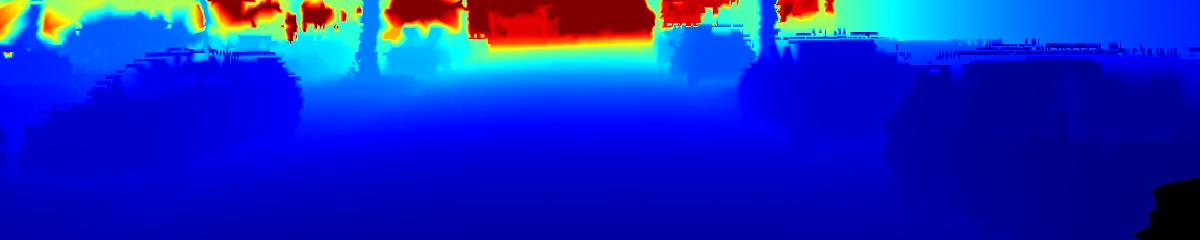} \vspace{-0.03em}&
  \includegraphics[width=0.16\textwidth]{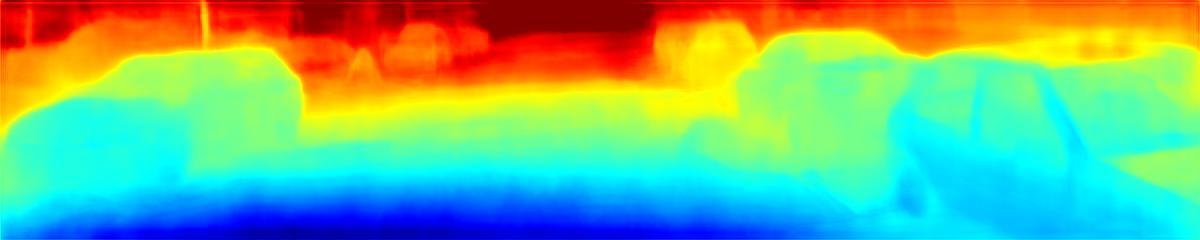} \vspace{-0.03em}&
  \includegraphics[width=0.16\textwidth]{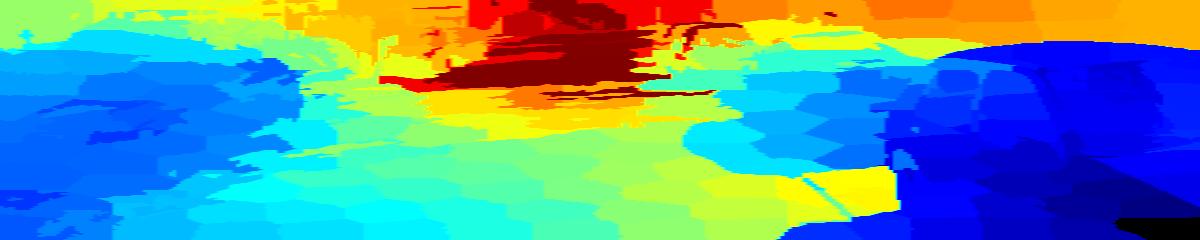} \vspace{-0.03em}&
  \includegraphics[width=0.16\textwidth]{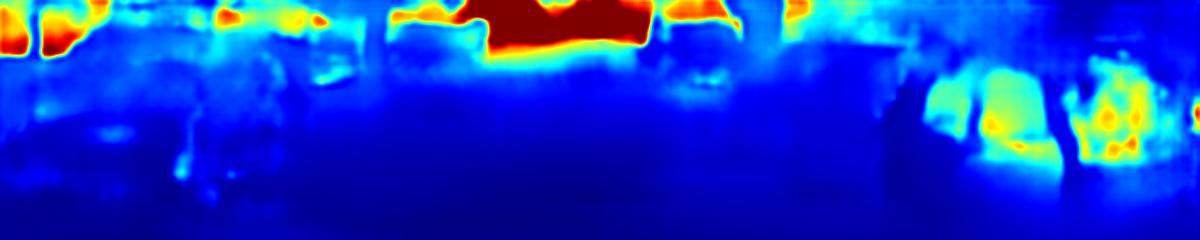} \vspace{-0.03em}&
  \includegraphics[width=0.16\textwidth]{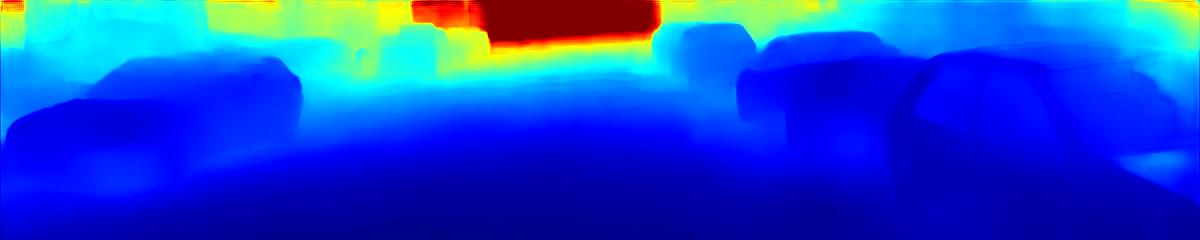} \vspace{-0.03em} \\
  \includegraphics[width=0.16\textwidth]{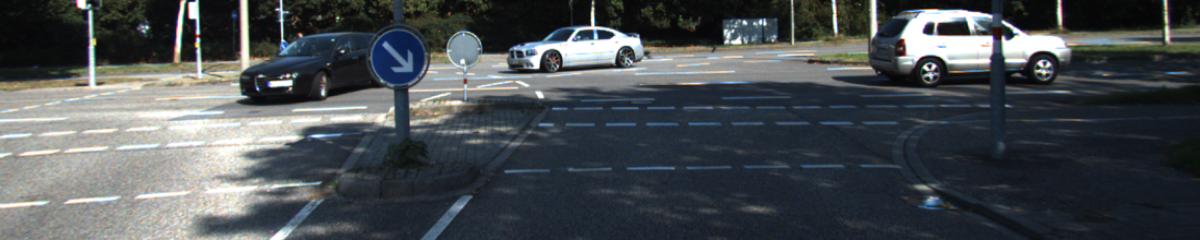} \vspace{-0.03em}&
  \includegraphics[width=0.16\textwidth]{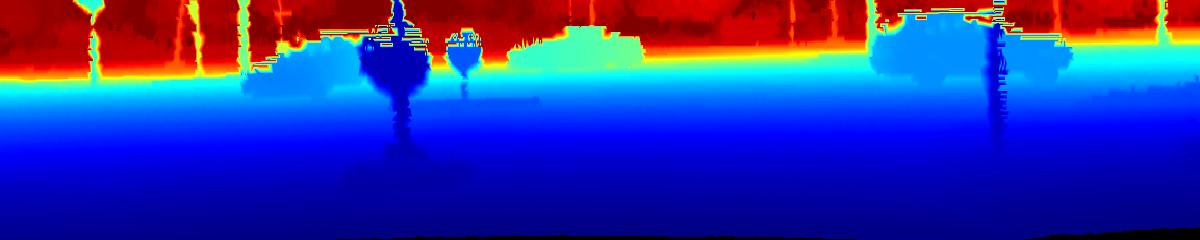} \vspace{-0.03em}&
  \includegraphics[width=0.16\textwidth]{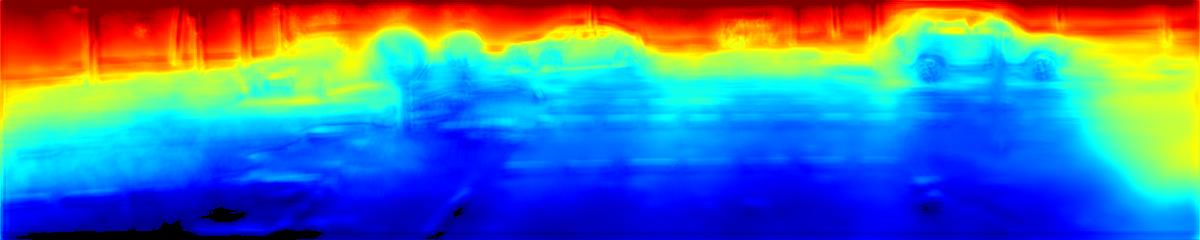} \vspace{-0.03em}&
  \includegraphics[width=0.16\textwidth]{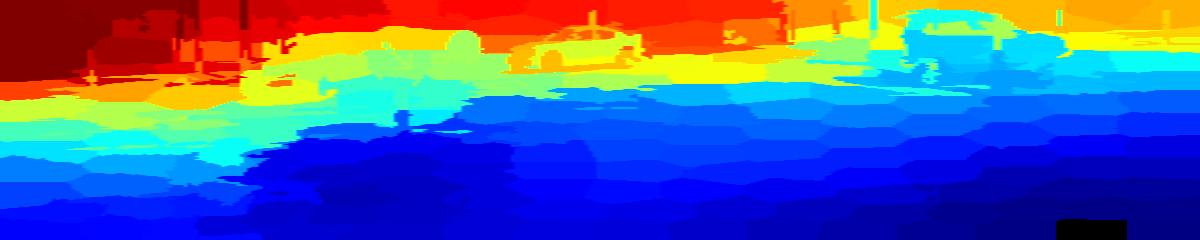} \vspace{-0.03em}&
  \includegraphics[width=0.16\textwidth]{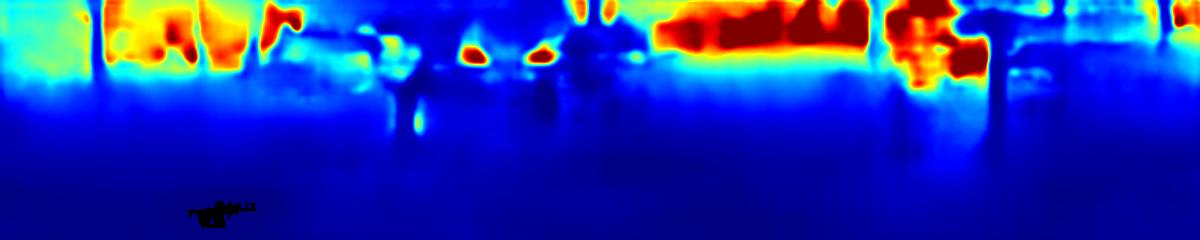} \vspace{-0.03em} &
  \includegraphics[width=0.16\textwidth]{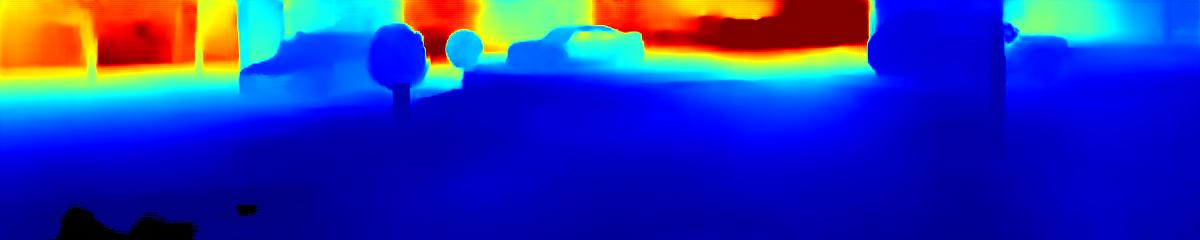} \vspace{-0.03em}\\
                  {\footnotesize (a) Image} \vspace{-0.1em}&
                  {\footnotesize (b) GT} \vspace{-0.1em}&
                  {\footnotesize \makecell{(c) DIW~\cite{chen2016single}}} \vspace{-0.1em}&
                  {\footnotesize \makecell{(d) Best NYU~\cite{Depth2015Liu}}} \vspace{-0.1em}&
                  {\footnotesize \makecell{(e) Best Make3D~\cite{laina2016deeper}}} \vspace{-0.1em} &
                  {\footnotesize (f) \DatasetShort} \vspace{-0.1em}
\end{tabular}
    \caption{\textbf{Depth predictions on KITTI.}  (Blue$=$near,
      red$=$far.) None of the models were trained on KITTI data. \label{fig:KITTI}} \vspace{-1.0em}
\end{figure*}

\begin{table}[t]
\centering
{\small
\begin{tabular}{llr}
  \toprule
Training set & Method & WHDR\% \\
\midrule
DIW & Chen~\etal~\cite{chen2016single} & 22.14 \\
\midrule
KITTI & Zhou~\etal~\cite{zhou2017unsupervised} & 31.24  \\
 & Godard~\etal~\cite{monodepth17} & 30.52 \\
\midrule
NYU & Eigen~\etal~\cite{eigen2015predicting} & 25.70 \\
 & Laina~\etal~\cite{laina2016deeper} & 45.30 \\
 & Liu~\etal~\cite{Depth2015CVPR} & 28.27  \\
\midrule
Make3D & Laina~\etal~\cite{laina2016deeper} & 31.65  \\
 & Liu~\etal~\cite{Depth2015CVPR} & 29.58 \\
\midrule
\DatasetShort & Ours & 24.55 \\
\bottomrule
\end{tabular}
}
\caption{{\bf Results on the DIW test set for various training
    datasets and approaches.} Columns are as in
  Table~\ref{tb:tb4}. \label{tb:tb6}}
\vspace{-0.5em}
\end{table}


\begin{figure}[t]
  \centering
    \begin{tabular}{@{\hspace{-0.1em}}c@{\hspace{-0.1em}}c@{\hspace{-0.1em}}c@{\hspace{-0.1em}}c@{\hspace{-0.1em}}c@{\hspace{-0.1em}}}
        \includegraphics[width=0.19\columnwidth]{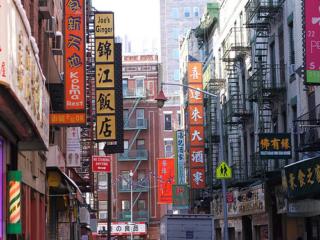} \vspace{-0.1em} & 
        \includegraphics[width=0.19\columnwidth]{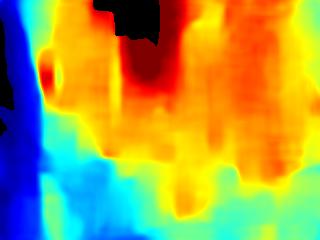}  \vspace{-0.1em} &
        \includegraphics[width=0.19\columnwidth]{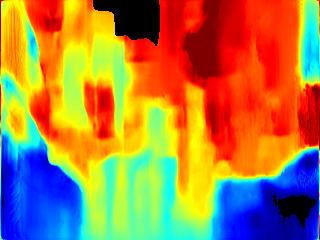}  \vspace{-0.1em} &
        \includegraphics[width=0.19\columnwidth]{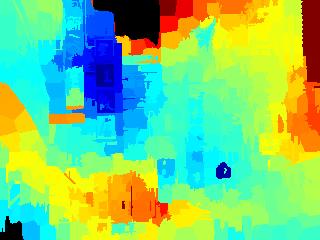}  \vspace{-0.1em} &
        \includegraphics[width=0.19\columnwidth]{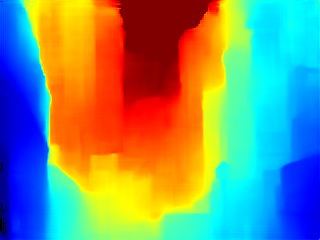}  \vspace{-0.1em} \\
        \includegraphics[width=0.19\columnwidth]{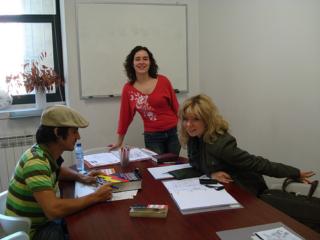} \vspace{-0.1em} & 
        \includegraphics[width=0.19\columnwidth]{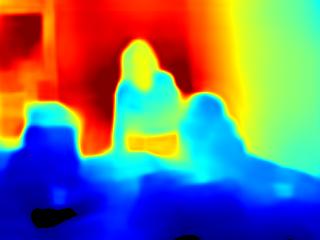}  \vspace{-0.1em} &
        \includegraphics[width=0.19\columnwidth]{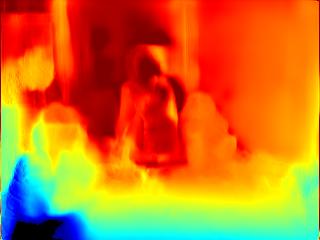}  \vspace{-0.1em} &
        \includegraphics[width=0.19\columnwidth]{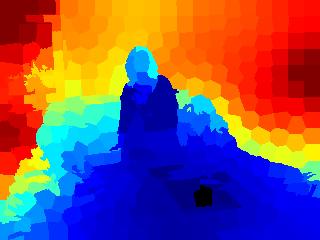}  \vspace{-0.1em} &
        \includegraphics[width=0.19\columnwidth]{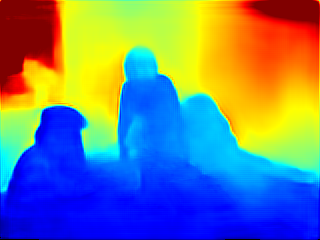}  \vspace{-0.1em} \\
        \includegraphics[width=0.19\columnwidth]{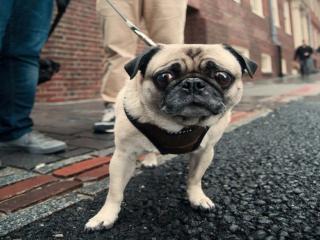} \vspace{-0.1em} & 
        \includegraphics[width=0.19\columnwidth]{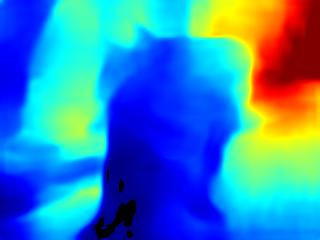}  \vspace{-0.1em} &
        \includegraphics[width=0.19\columnwidth]{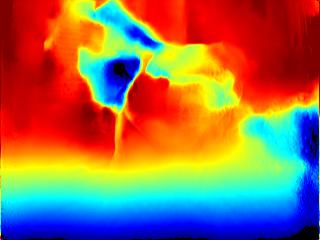}  \vspace{-0.1em} &
        \includegraphics[width=0.19\columnwidth]{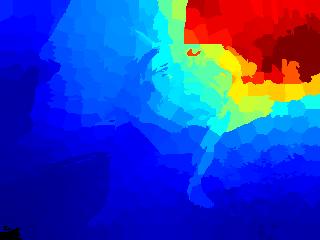}  \vspace{-0.1em} &
        \includegraphics[width=0.19\columnwidth]{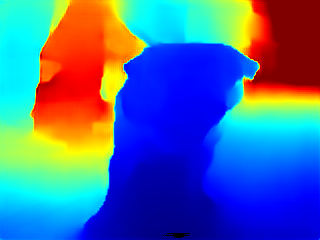}  \vspace{-0.1em} \\
       {\scriptsize (a) Image} & {\scriptsize (b) NYU~\cite{eigen2015predicting}} & {\scriptsize (c) KITTI~\cite{monodepth17}} & {\scriptsize (d) Make3D~\cite{Depth2015CVPR}} & {\scriptsize (e) Ours} \vspace{-0.5em}
    \end{tabular} 
  \caption{ \textbf{Depth predictions on the DIW test set.}
    (Blue$=$near, red$=$far.) Captions are described in Figure~\ref{fig:KITTI}. None of the models were trained on DIW
    data.\label{fig:DIW_test}} \vspace{-0.5em}    
\end{figure}

\smallskip
\noindent\textbf{Make3D.}
To test on Make3D, we follow the protocol of \cite{Depth2015Liu, laina2016deeper}, ,resizing all images to 345$\times$460, and removing ground truth depths larger than 70m (since Make3D
data is unreliable at large distances). We train our
network only on \DatasetShort using our full loss.
Table~\ref{tb:tb4} shows numerical results, including comparisons to
several methods trained on both Make3D and non-Make3D data, and Figure~\ref{fig:Make3d} visualizes depth predictions from our model
and several other non-Make3D-trained models. Our
network trained on \DatasetShort have the best
performance among all non-Make3D-trained models. Finally, the last row of
Table~\ref{tb:tb4} shows that our model fine-tuned on Make3D achieves
better performance than the state-of-the-art.


\smallskip
\noindent\textbf{KITTI.} Next, we evaluate our model on the KITTI test
set based on the split of~\cite{eigen2014depth}. As with our Make3D
experiments, we do not use images from KITTI during training. The
KITTI dataset is very different from ours, consisting of driving
sequences that include objects, such as sidewalks, cars, and people,
that are difficult to reconstruct with SfM/MVS.
Nevertheless, as shown in Table~\ref{tb:tb5}, our
\DatasetShort-trained network still outperform approaches trained on
non-KITTI datasets. 
Finally, the last row of Table~\ref{tb:tb5} shows that we can achieve
state-of-the-art performance by fine-tuning our network on KITTI
training data.
Figure~\ref{fig:KITTI} shows visual comparisons between our results
and models trained on other non-KITTI datasets. One can see that we
achieve much better visual quality compared to other non-KITTI
datasets, and our predictions can reasonably capture nearby objects
such as traffic signs, cars, and trees, due to our ordinal depth loss.


\smallskip
\noindent\textbf{DIW.} Finally, we test our network on the DIW
dataset~\cite{chen2016single}. DIW consists of Internet photos with
general scene structures. Each image in DIW has a single pair of
points with a human-labeled ordinal depth relationship. As with Make3D
and KITTI, we do not use DIW data during training. For DIW, quality is
computed via the {\em Weighted Human Disagreement Rate} (WHDR),
which measures the frequency of disagreement between predicted depth
maps and human annotations on a test set. Numerical results are shown
in Table~\ref{tb:tb6}. Our \DatasetShort-trained network again has the
best performance among all non-DIW trained models.
Figure~\ref{fig:DIW_test} visualizes our predictions and those of
other non-DIW-trained networks on DIW test images. Our predictions
achieve visually better depth relationships.
Our method even works reasonably well for challenging scenes such as
offices and close-ups.

\section{Conclusion}
We presented a new use for Internet-derived SfM+MVS data: generating
large amounts of training data for single-view depth prediction. We
demonstrated that this data can be used to predict state-of-the-art
depth maps for locations never observed during training, and
generalizes very well to other datasets.
However, our method also has a number of limitations. MVS methods
still do not perfectly reconstruct even static scenes, particularly
when there are oblique surfaces (e.g., ground), thin or complex
objects (e.g., lampposts), and difficult materials (e.g., shiny
glass). Our method does not predict metric depth; future work in SfM
could use learning or semantic information to correctly scale scenes.
Our dataset is currently biased towards outdoor landmarks, though by
scaling to much larger input photo collections we will find more
diverse scenes.
Despite these limitations, our work points towards the Internet as an
intriguing, useful source of data for geometric learning problems.

\smallskip
{\small 
\noindent \textbf{Acknowledgments.} We thank the anonymous reviewers
for their valuable comments. This work was funded by the National
Science Foundation under grant IIS-1149393.
}

\newpage
{\small
\bibliographystyle{ieee}
\bibliography{paper}
}

\end{document}